\documentclass[preprint,12pt,num]{elsarticle}

\usepackage{amssymb}
\usepackage{amsmath}

\usepackage{todonotes}
\usepackage{graphicx}
\usepackage{hyperref}
\usepackage{todonotes}
\usepackage{pgfplots}
\usepackage{subcaption}
\usepackage{sidecap}
\usepackage{booktabs}
\pgfplotsset{compat=1.18}
\usepackage{float}
\usepackage{array}

\usepackage{multirow}
\usepackage{tabularx}

\usepackage{makecell}

\usepackage{xcolor}
\usepackage{soul} 
\usepackage{marginnote}

\hypersetup{hidelinks}
\reversemarginpar 

\usepackage{xcolor}
\usepackage{soul}        
\usepackage{todonotes}  

\reversemarginpar

\journal{Solar Energy}

\begin{document}

\begin{frontmatter}




\title{
A Generative Approach for Improving Multi-Label Defect Classification in Photovoltaic Modules
}

%

\author{Abdul Mueez}

\author{Yogesh S. Rawat}

\author{Shruti Vyas\corref{cor1}}
\ead{shruti@ucf.edu}

\cortext[cor1]{Corresponding author.}

\nonumnote{%
© 2026. This manuscript version is made available under the
CC BY-NC-ND 4.0 license
(\url{https://creativecommons.org/licenses/by-nc-nd/4.0/}).
Published version:
\url{https://doi.org/10.1016/j.solener.2026.114943}.%
}

\address{Institute of Artificial Intelligence, University of Central Florida,
4356 Scorpius St., Orlando, FL 32816, United States of America}



\begin{abstract}

This paper addresses the challenge of multi-label defect classification in electroluminescence (EL) images of photovoltaic (PV) cells. Training models on images where multiple defects co-occur creates learning ambiguity, making it difficult to disentangle visual features for specific defect types, a problem compounded by the scarcity of examples for individual classes. To tackle this, we introduce Generative Defect Isolation (GDI), utilizing the LaMa inpainting model with Fast Fourier Convolutions to remove selected defects and generate realistic, single-defect training samples. Extensive experiments on Vision Transformer (ViT-S, ViT-L) and EfficientNetV2-L architectures demonstrate that GDI significantly outperforms baselines. The performance gains are most pronounced in low-data scenarios; class-wise analysis shows substantial improvements, boosting the F1-Score for rare defect classes by up to 63.6\%. Furthermore, GDI effectively resolves learning ambiguity from co-occurring defects, yielding a 26\% reduction in such co-occurring classification errors. Our work establishes GDI as an effective method for maximizing the value of existing segmentation datasets and sets a new performance benchmark for multi-label classification in this domain.

\end{abstract}


\end{frontmatter}




\section{Introduction}
\label{sec:intro}

The global transition towards renewable energy sources has positioned photovoltaics (PV) as a cornerstone of sustainable energy production, with global installed capacity growing exponentially \citep{hijjawi2023review, demirci2024improved}. As the scale of PV plants increases, particularly in complex operating environments, the industry is increasingly shifting toward intelligent operation and maintenance (O\&M) strategies to ensure system safety and economic efficiency \citep{tang2024module}. However, the efficacy of these intelligent inspection systems is highly susceptible to a variety of defects that can arise during manufacturing, transportation, or from environmental stressors in the field \citep{hijjawi2023review, chindarkkar2020deep}. These defects, ranging from micro-cracks to cell interconnect failures, can lead to significant power degradation, reduced module lifetime and potential safety hazards \citep{kontges2014review, chindarkkar2020deep}.

Among various non-destructive testing (NDT) techniques, Electroluminescence (EL) imaging has emerged as a highly effective method for quality control in the PV industry \citep{fuyuki2009photographic, hijjawi2023review}. EL imaging provides high-resolution spatial information that can reveal even the finest defects, such as micro-cracks, which are often invisible to the naked eye or other inspection methods like infrared thermography \citep{deitsch2019automatic, fioresi2022automated}. However, the manual inspection of the vast number of EL images generated during high-throughput manufacturing and large-scale field testing is a major bottleneck. This process is time-consuming, expensive and prone to subjective errors and inconsistencies by human operators \citep{deitsch2019automatic, fioresi2022automated}.

To overcome these limitations, the research community has increasingly focused on developing automated defect detection systems. Driven by advances in computer vision, deep learning - particularly Convolutional Neural Networks (CNNs) - has become the predominant approach, consistently demonstrating superior performance over traditional image processing techniques \citep{hijjawi2023review, su2023pvel}.

Despite these advancements, several key challenges persist. A significant portion of existing deep learning models is designed for binary classification (i.e., distinguishing between ``defective'' and ``non-defective'' cells) \citep{deitsch2019automatic, akram2019cnn}. In practical scenarios, however, a single PV cell often exhibits multiple, co-occurring defects. A recent comprehensive review highlights this as a critical limitation in current research: existing defect detection solutions predominantly focus on identifying single defect types per image, failing to address the reality that multiple defect types frequently coexist \citep{tang2024module}. Consequently, the development of algorithms capable of handling this multi-label complexity has been identified as a key future direction for intelligent PV maintenance \citep{tang2024module}. Furthermore, the performance of deep learning models is heavily reliant on the availability of large, diverse and accurately annotated datasets, which remain scarce in this specialized domain \citep{demirci2024improved}.

The inherent class imbalance found in real-world data, where non-defective cells vastly outnumber any single class of defect, further complicates the training of robust and generalizable models \citep{fioresi2022automated, demirci2024improved}.

While generative methods have been used to create new images from scratch, a key opportunity remains untapped: repurposing the rich, pixel-level detail from existing segmentation datasets to resolve ambiguity in the more common task of multi-label classification.


To address these challenges, this paper presents a data-centric approach to enhance multi-label defect classification in PV cell EL images. We utilize the publicly available UCF-EL-Defect dataset \citep{fioresi2022automated}, which provides rich pixel-level segmentation annotations for nine distinct defect classes. We first reframe this segmentation dataset for a multi-label classification task. Our main contribution is the introduction of an annotation-guided data augmentation strategy we term Generative Defect Isolation (GDI). This technique leverages the ground-truth segmentation masks to computationally ``inpaint'' or remove selected defects from an image. By doing so, GDI generates new, high-quality training examples of cells with isolated, single defects from source images that originally contained multiple, overlapping defects. This method directly targets the problem of data scarcity for underrepresented defect combinations and helps the model learn to disentangle features associated with different defect types.

The main contributions of this work are:

\begin{enumerate}

\item We propose Generative Defect Isolation (GDI), a technique that utilizes the LaMa inpainting architecture \citep{suvorov2022resolution} to leverage segmentation annotations. This allows us to iteratively isolate defects, creating new, realistic training samples from multi-defect images.
\item We demonstrate through extensive experiments using Vision Transformer (ViT-S, ViT-L) and EfficientNetV2-L architectures that models trained with GDI significantly outperform a baseline on a challenging multi-label defect classification task.
\item By leveraging GDI, we establish a new performance benchmark for multi-label classification on the UCF-EL-Defect dataset, providing a valuable resource for future research.
\end{enumerate}

The remainder of this paper is organized as follows: Section \ref{sec:related_work} reviews the existing literature on automated defect detection in EL images. Section \ref{sec:methodology} details our proposed methodology, including the dataset preparation and the annotation-guided augmentation technique. Section \ref{sec:results} presents the experimental setup and results, followed by a discussion in Section \ref{sec:discussion}. Finally, Section \ref{sec:conclusion} concludes the paper.

\section{Related Work}
\label{sec:related_work}
The automated analysis of EL images for PV quality control has been an active area of research, evolving from traditional image processing to sophisticated deep learning architectures. This section reviews key developments in this field, with a focus on methodologies for defect classification and localization and the various strategies employed to address prevalent data-related challenges.

\subsection{Automated Defect Classification and Localization}


Defect detection techniques for PV modules are generally categorized into electrical parameter analysis (I/V curves) and imaging techniques (Visible, IR, EL/PL) \citep{tang2024module}. While I/V curve analysis is cost-effective, it is limited by its inability to precisely locate defects or detect tiny anomalies \citep{tang2024module}. Similarly, while UAV-based visible and IR inspections offer high efficiency, they often lack the ``inner perspective'' required to identify detailed cell-level faults \citep{tang2024module}. Electroluminescence (EL) imaging has thus emerged as a superior method for detailed defect diagnosis, offering high-resolution insight into the module's internal state.


Early research into automated EL analysis relied on classical computer vision and machine learning techniques. For instance, \cite{tsai2012defect} used Fourier image reconstruction to detect cracks and breaks, while other work from the same group used Independent Component Analysis (ICA) to create basis images for identifying defects \citep{tsai2013defect}. Other approaches have also included Support Vector Machines (SVMs) trained on hand-crafted features \citep{deitsch2019automatic}. While effective to a degree, these methods often struggled with the high variability of defect appearances and the complex textures of polycrystalline cells.

With the rise of deep learning, CNNs have become the standard for defect detection tasks due to their ability to automatically learn hierarchical features from raw pixel data. A large body of work has demonstrated the success of CNNs for binary classification of PV cells as either ``defective'' or ``functional'' \citep{deitsch2019automatic, akram2019cnn, chindarkkar2020deep}. Various architectures, from custom lightweight networks \citep{akram2019cnn, zhang2024lightweight} to established backbones like VGG \citep{deitsch2019automatic, demirci2024improved} and ResNet \citep{fan2022automatic}, have been successfully applied. More specialized architectures, such as high-resolution networks designed to preserve fine-grained details \citep{zhao2023hrnet} and methods that fuse multi-channel CNNs to capture diverse features \citep{zhang2020detection}, have also been developed to improve detection accuracy.

Beyond simple classification, some research has focused on providing more granular information about the defect's location and extent. This is typically achieved through either object detection, which places a bounding box around a defect, or semantic segmentation, which assigns a class label to every pixel in the image. \cite{su2023pvel} introduced PVEL-AD, a large-scale dataset with bounding box annotations for ten different anomaly types and benchmarked several object detection models. Providing the most detailed level of annotation, \cite{fioresi2022automated} developed the UCF-EL-Defect dataset, which our work is based on, for semantic segmentation of nine defect classes using a DeepLabv3+ model. Other works have also explored segmentation, such as \cite{zhao2021deep} who used Mask R-CNN for localizing 14 defect types in a production line setting. Our work sits at the intersection of these approaches, leveraging the rich annotations of a segmentation dataset to improve the performance of a classification task.

\subsection{Data Augmentation and Imbalance Handling}
A persistent challenge in applying deep learning to PV defect detection is the scarcity of large, labeled datasets and the inherent class imbalance of real-world data \citep{hijjawi2023review, demirci2024improved}. Consequently, data augmentation has become a critical component of the training pipeline.

Standard geometric and color-space transformations, such as random flipping, rotation and brightness adjustments, are widely used to increase the diversity of the training set and improve model robustness \citep{bartler2018automated, zhang2024lightweight, demirci2024improved}. However, the effectiveness of these simple transformations can be limited.


Prior work has explored using generative models, particularly Generative Adversarial Networks (GANs), to generate synthetic data by synthesizing new artificial EL images of defective cells \citep{demirci2024improved, ebied2025advanced}. A significant limitation of GAN-based synthetic data generation is highly hardware-demanding and prone to training instability, which can lead to convergence failure and the production of poor-quality data \cite{demirci2024improved}. Additionally, the generative process relies on global image creation from noise, which may result in mode collapse where the generator fails to produce a diverse range of random patterns for the discriminator \cite{demirci2024improved}. Notably, \citep{demirci2024improved} formulate the problem as a binary classification task, while \citep{ebied2025advanced}, despite proposing a multi-class data generation strategy, ultimately evaluate their approach under a binary classification setting. Such evaluations overlook the fundamentally different and significantly more challenging nature of multi-label classification, where multiple defect types can co-occur within a single sample.

As an alternative to purely generative approaches, some have used physics-based simulations to create realistic defect images. \cite{fioresi2022automated}, in the creation of the UCF-EL-Defect dataset, supplemented their real-world images with 256 physically realistic simulated images to address the severe class imbalance of interconnect and contact defects.

The existing literature showcases a clear trend towards using advanced data augmentation to overcome dataset limitations. While methods like GANs generate entirely new images and geometric transforms alter existing ones, our proposed work introduces a distinct, hybrid approach. By leveraging precise segmentation annotations for inpainting, we do not create wholly synthetic data but rather \textit{disentangle} real, co-occurring defects from a single image. This methodology creates a unique dataset of semi-synthetic, high-quality images tailored specifically for the challenging multi-label classification problem, bridging the gap between classification-focused and segmentation-focused research.

More broadly, an alternative strategy to address data scarcity involves semi-supervised learning \citep{jha2025advancing}, where models are trained on less expensive annotations like bounding boxes or image-level labels. While such an approach aims to lower the initial cost of data collection, our work addresses a different scenario: leveraging and maximizing the value of existing, high-density segmentation annotations to improve performance on a related downstream task.

\section{Methodology}
\label{sec:methodology}

Our approach expands a dataset containing complex, co-occurring defects by supplementing it with simplified, single-defect images, creating a larger and more robust collection for training a multi-label classifier. This is achieved through a multi-stage process we call Generative Defect Isolation (GDI). This section details the initial dataset, the preprocessing steps and the core GDI pipeline.

\subsection{Dataset and Initial Processing}
\label{sec:methodology_dataset_and_init_proc}

The foundation of our work is the publicly available UCF-EL-Defect dataset \citep{fioresi2022automated}. As provided by the original authors, the dataset consists of isolated, cell-level images. These exhibit significant resolution variance—ranging from $165 \times 169$ to $1174 \times 1144$ pixels—across 491 unique sizes. This variance introduces a degree of visual ambiguity; in lower-resolution samples, distinguishing between fine-grained defect types can be challenging even for human experts.

Importantly, the dataset provides detailed, pixel-level segmentation annotations for nine distinct defect classes. Although originally designed for segmentation, we apply the dataset to the more common industrial task of multi-label classification. The dataset annotations contain two additional classes, \textit{Contact\allowbreak\_BeltMarks} and \textit{Unknown}, which were not utilized in the original study. We chose to incorporate these two classes in our analysis. Furthermore, to build a complete classification scheme, we introduced a \textit{No\allowbreak\_Defect} class for images confirmed to be defect-free. An image is assigned multiple labels if it contains annotations for more than one defect type. This process results in the following 12 target classes for our task: \textit{Contact\allowbreak\_BeltMarks, Contact\allowbreak\_Corrosion, Contact\allowbreak\_FrontGridInterruption, Contact\allowbreak\_NearSolderPad, Crack\allowbreak\_Closed, Crack\allowbreak\_Isolated, Crack\allowbreak\_Resistive, Interconnect\allowbreak\_BrightSpot, Interconnect\allowbreak\_Disconnected, Interconnect\allowbreak\_HighlyResistive, No\allowbreak\_Defect, Unknown}.

\subsection{Generative Defect Isolation (GDI)}

\begin{figure}[H]
    \centering
    \includegraphics[width=\textwidth]{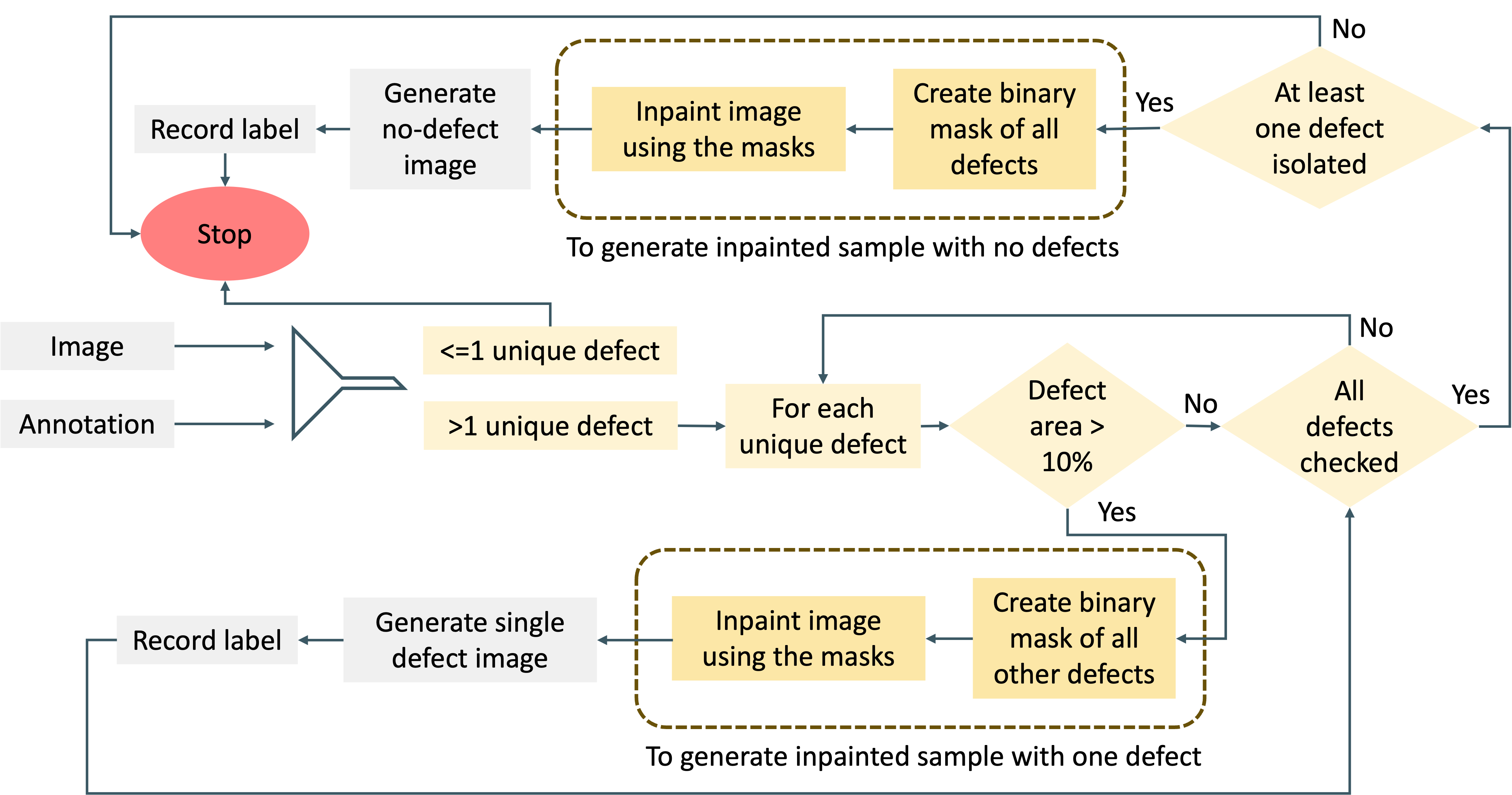} 
    \caption{\textbf{\textit{The proposed Generative Defect Isolation (GDI) process:}} For images with multiple defects, the method first iteratively isolates each defect by inpainting all others. After this loop, it generates a final ``no-defect'' sample by creating a combined mask of all defects and inpainting them simultaneously. This creates multiple single-defect samples and one defect-free sample from a single source image.}
    \label{fig:dia_flowchart}
\end{figure}

The central contribution of this work is the GDI pipeline, designed to resolve the learning ambiguity that arises when a model is trained on images containing multiple, overlapping defect types. The GDI method proceeds in three algorithmic steps for each candidate image: (1) Defect Selection and Filtering, (2) Annotation-Guided Mask Generation and (3) Neural Inpainting. By utilizing ground-truth segmentation masks, this process computationally removes selected defects to generate new, high-quality images that exhibit only a single, isolated defect or no defects. The entire process is illustrated schematically in Figure \ref{fig:dia_flowchart}.

\subsubsection{Defect Selection and Filtering}
The GDI process is applied exclusively to images that contain more than one unique defect type. For such an image, we iterate through each of its annotated defects to potentially isolate it. However, to ensure that the resulting augmented image contains a visually significant defect rather than a minor blemish that could be mistaken for a defect-free cell, we apply an area-based filter. As shown in Figure~\ref{fig:dia_flowchart}, a specific defect type within an image is only considered for isolation if its total annotated area covers more than 10\% of the entire image area. This prevents the generation of trivial samples and focuses the augmentation effort on prominent defects. Further, if at least one defect satisfies this criterion and is successfully isolated, we additionally create a combined mask covering all annotated defects in the source image and inpaint these regions to generate a corresponding defect free sample.

\subsubsection{Annotation-Guided Mask Generation}
For a multi-defect image and a chosen target defect class, $D_{target}$, that meets the area criteria, the next step is to generate a mask to remove all other defects. A binary mask, $M_{other}$, is created by rasterizing the geometric annotations (polygons, circles, ellipses, rectangles) of all non-target defect classes present in the image. This mask precisely covers all pixels belonging to the defects slated for removal.

To prevent artifacts at the boundary of the inpainted region, this initial mask is dilated. Dilation expands the masked area, creating a small buffer zone around the unwanted defects. This ensures that the inpainting algorithm seamlessly blends the synthesized texture with the surrounding original image content. A square dilation kernel of size $15 \times 15$ was used. This value was selected through qualitative inspection as it provided the best balance between fully masking defect boundaries and preserving surrounding textures, thereby minimizing inpainting artifacts.

\subsubsection{Neural Inpainting with LaMa}

\begin{figure}[htbp]
    \centering
    \includegraphics[width=\textwidth]{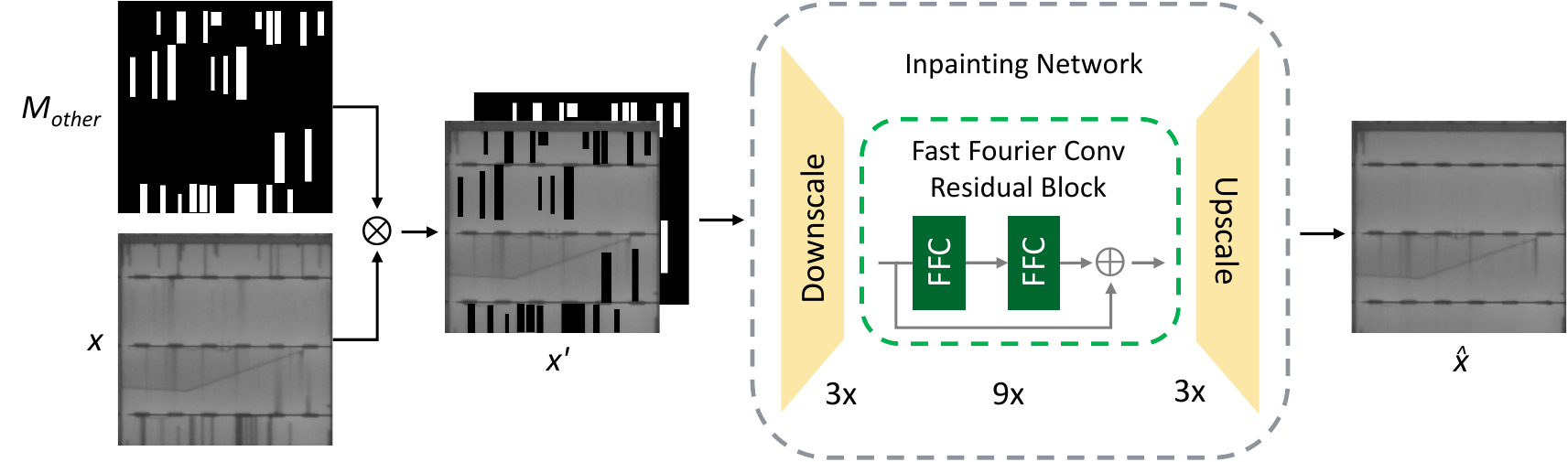}
    \caption{\textbf{\textit{The proposed inpainting workflow:}} The original image ($x$) and the binary mask of other defects ($M_{other}$) are combined to create a masked input tensor ($x'$). This tensor is processed by an inpainting network featuring an encoder-decoder architecture: three downscale blocks reduce the spatial resolution, a core of nine Fast Fourier Convolution (FFC) residual blocks processes the features and three upscale blocks reconstruct the final restored image ($\hat{x}$). This approach is based on the LaMa \citep{suvorov2022resolution} model.}
    \label{fig:inpainting_process}
\end{figure}

With the dilated mask prepared, we employ the LaMa (Large Mask Inpainting) model \citep{suvorov2022resolution} to fill the masked regions. We utilize the implementation provided by the ``Inpaint Anything'' codebase \citep{yu2023inpaint}. This architecture was specifically selected for its ability to handle high-resolution inputs with complex periodic structures, replacing irregular masked regions with coherent, realistic textures \citep{suvorov2022resolution}. Unlike standard convolutional methods that often fail to maintain global structural consistency across large masks, the Fast Fourier Convolutions in LaMa enable the precise reconstruction of the periodic grid lines inherent to PV cells. The workflow is illustrated in Figure \ref{fig:inpainting_process} and the process consists of the following technical steps:

\begin{enumerate}
    \item \textbf{Input Tensor Construction:} For a source image $x$ and the generated binary mask $M_{other}$ (where 1 denotes a pixel to inpaint), we first compute the masked image $x \odot (1 - M_{other})$. We then construct a four-channel input tensor $x'$ by stacking the masked image and the mask itself: $x' = \text{stack}(x \odot (1 - M_{other}), M_{other})$.
    
    \item \textbf{Feature Encoding and Fast Fourier Convolutions:} The tensor $x'$ is processed by a feed-forward network with a ResNet-like topology. The encoder utilizes three downsampling blocks to reduce the spatial resolution, followed by a bottleneck of nine Fast Fourier Convolution (FFC) residual blocks. The FFC blocks are the core algorithmic component; unlike standard convolutions, they split the feature channels into two parallel branches:
    \begin{itemize}
        \item \textbf{Local Branch:} Uses conventional convolutions to capture local texture details.
        \item \textbf{Global Branch:} Applies a Real Fast Fourier Transform (Real FFT2d) to transition features into the frequency domain, performs a $1 \times 1$ convolution to capture global context and recovers the spatial structure via an Inverse Real FFT2d.
    \end{itemize}
    This spectral transform allows the network to learn global interactions and periodic patterns, specifically the grid lines and busbars inherent to PV cells at any depth, effectively providing an image-wide receptive field.

    \item \textbf{Reconstruction and Loss:} Finally, three upsampling blocks reconstruct the feature maps to the original resolution, yielding the restored image $\hat{x}$. The network is trained using a combined objective function comprising a High Receptive Field (HRF) perceptual loss and an adversarial loss. The HRF perceptual loss utilizes a pre-trained segmentation network with dilated convolutions to enforce structural consistency between the inpainted region and the surrounding valid pixels, while the adversarial loss, calculated via a discriminator operating on local patches, ensures the synthesized textures (e.g., silicon grain) are photorealistic \citep{suvorov2022resolution}.
\end{enumerate}


The augmented images were generated on a system equipped with an Intel Xeon W-3335 CPU and an NVIDIA RTX A5500 GPU. The GDI preprocessing step required approximately 3.49 seconds per image. It is important to note that the entire GDI pipeline, including the use of the LaMa model, is a one-time, offline pre-processing step used to augment the training dataset. It carries no computational overhead at inference time, thus preserving the efficiency of the final classification model.

\subsubsection{Augmented Dataset Creation}
This isolation process is repeated for every defect type in the original image that passes the 10\% area filter. Consequently, a single source image with three significant, distinct defects can generate four training samples, one for each isolated defect and a defect free image.

After processing the entire source dataset, the newly generated single-defect images are combined with the original multi-defect training set to form the final, enhanced dataset. This hybrid approach ensures the model learns from both idealized defect patterns and the complex co-occurrences found in real-world scenarios. A corresponding CSV label file is generated for the new samples, where each image is mapped to a single positive class label (one-hot encoded) for the preserved defect.

\subsubsection{Dataset Composition}
\label{sec:dataset_composition}
To prepare our datasets for training and evaluation, the original UCF-EL-Defect dataset was first partitioned into an 80:20 training and test split. This split was performed on the source images before any augmentation was applied. Our Generative Defect Isolation (GDI) process was then applied exclusively to the training set to generate new, single-defect samples. The test set was reserved solely for evaluation and consists only of original, unaugmented images to ensure a fair assessment of the model's generalization capabilities.

Table \ref{tab:dataset_composition} provides a comprehensive overview of the final composition of both the training and test sets. The training set is composed of the original images from the 80\% split and the new samples generated by GDI, highlighting how the augmentation effectively supplements the base data across various classes. The test set consists entirely of the original images from the 20\% split. As detailed in Table \ref{tab:dataset_composition}, the final training set is constructed by combining the original images from the 80\% split with the new single-defect inpainted samples generated by GDI.

\begin{table}[h!]
\centering
\caption{\textbf{\textit{The final composition of the training and test datasets used in our experiments.}} The training set includes both original and GDI-generated samples, while the test set is composed exclusively of original images to ensure an unbiased evaluation. Note that as this is a multi-label dataset, an image can have multiple defects; therefore, the  `Overall Total' represents the number of unique images, not the sum of the defect instances in the columns.}
\label{tab:dataset_composition}
\resizebox{0.9\textwidth}{!}{%
\begin{tabular}{lrrrrr}
\toprule
\textbf{Class Name} & \multicolumn{3}{c}{\textbf{Training}} & \multicolumn{1}{c}{\textbf{Test}} \\
\cmidrule(lr){2-4} \cmidrule(lr){5-5}
& \textbf{Original} & \textbf{Inpainted} & \textbf{Total} & \textbf{Total} \\
\midrule
Contact\_BeltMarks & 9 & 5 & 14 & 5 \\
Contact\_Corrosion & 55 & 7 & 62 & 12 \\
Contact\_FrontGridInterruption & 6980 & 33 & 7013 & 1712 \\
Contact\_NearSolderPad & 4464 & 162 & 4626 & 1118 \\
Crack\_Closed & 2126 & 50 & 2176 & 514 \\
Crack\_Isolated & 1422 & 328 & 1750 & 333 \\
Crack\_Resistive & 2470 & 1475 & 3945 & 617 \\
Interconnect\_BrightSpot & 203 & 91 & 294 & 44 \\
Interconnect\_Disconnected & 161 & 5 & 166 & 69 \\
Interconnect\_HighlyResistive & 203 & 16 & 219 & 44 \\
No\_Defect & 598 & 1752 & 2350 & 257 \\
Unknown & 71 & 0 & 71 & 12 \\
\midrule
\textbf{Overall Total} & \textbf{10086} & \textbf{3924} & \textbf{14010} & \textbf{2620} \\
\bottomrule
\end{tabular}%
}
\end{table}

\section{Results}
\label{sec:results}

\subsection{Qualitative Analysis of Inpainting Quality}
\label{sec:inpainting_analysis}

\begin{figure}[htbp]
    \centering
    \captionsetup[subfigure]{font=scriptsize, skip=1pt}
    
    \begin{subfigure}{0.24\linewidth}
        \centering
        \includegraphics[width=\linewidth]{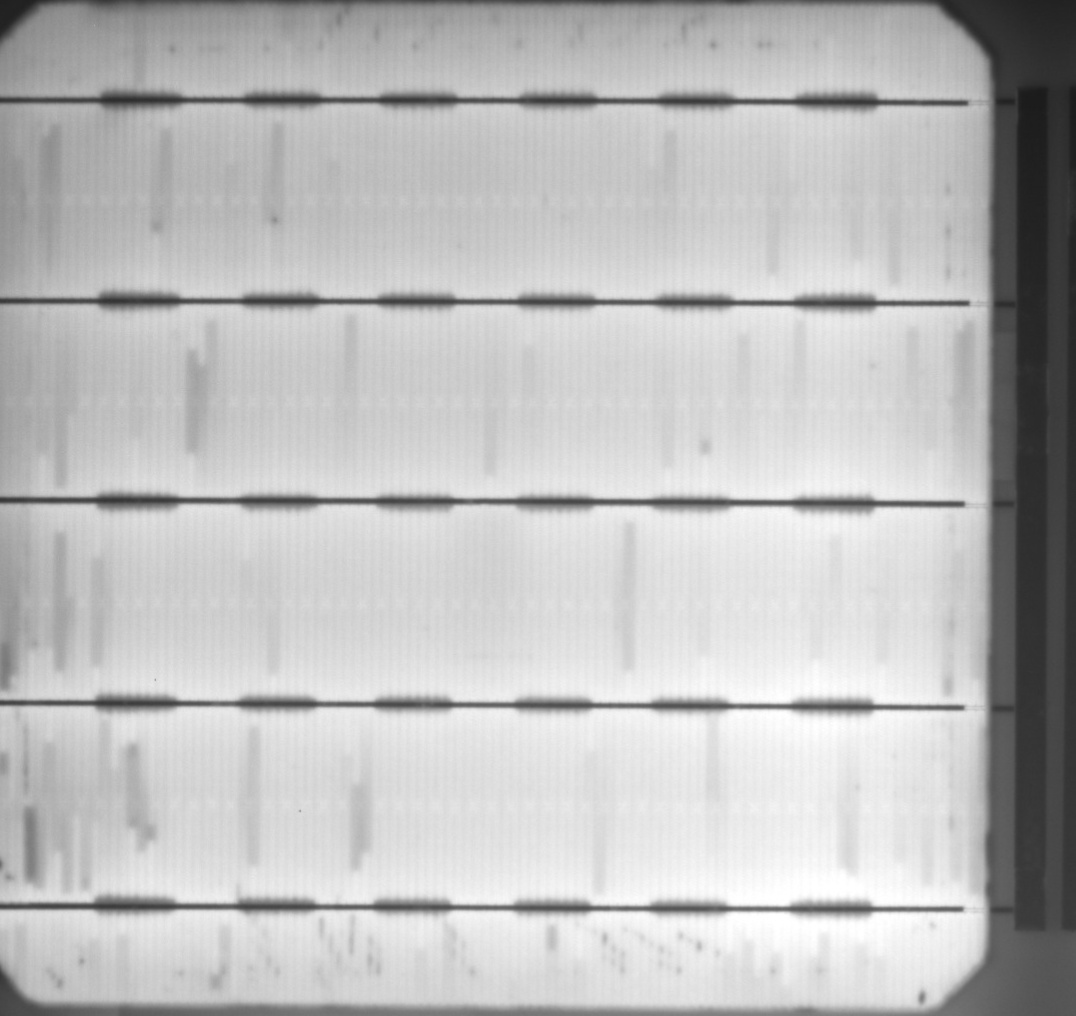}
        \caption[]{Original}
        \label{fig:1a_og}
    \end{subfigure}
    \begin{subfigure}{0.24\linewidth}
        \centering
        \includegraphics[width=\linewidth]{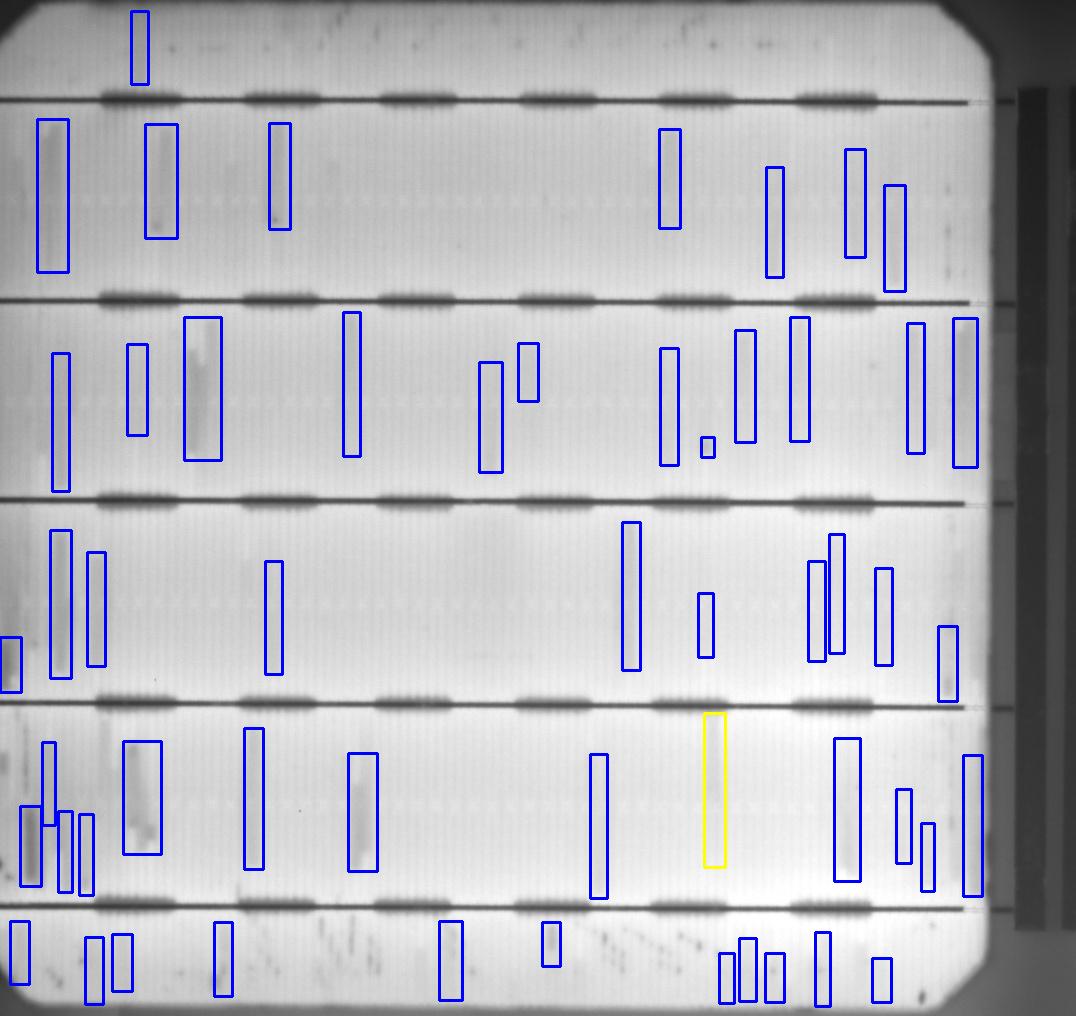}
        \caption[]{Boundaries}
        \label{fig:1b_bound}
    \end{subfigure}
    \begin{subfigure}{0.24\linewidth}
        \centering
        \includegraphics[width=\linewidth]{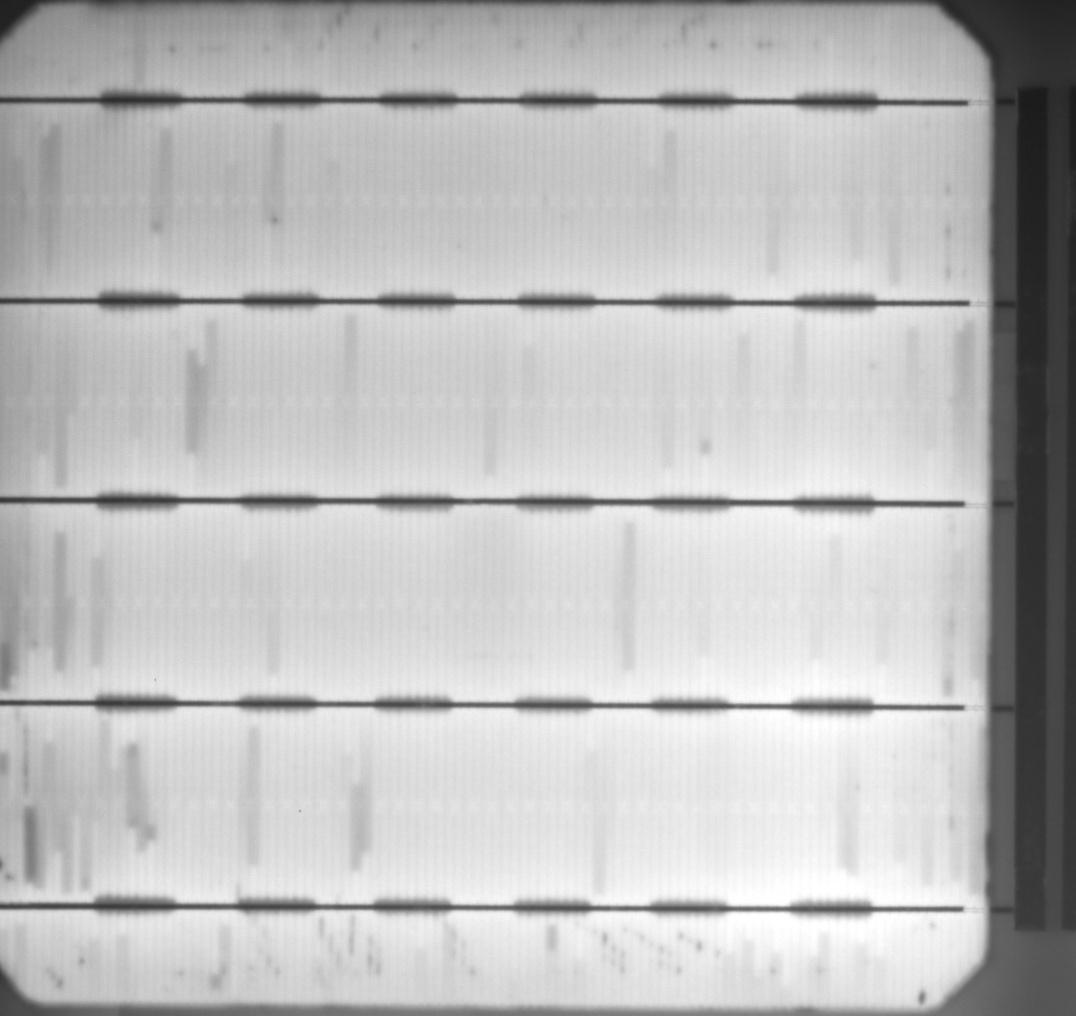}
        \caption[]{\textit{C\_FGInterruption}}
        \label{fig:1c_Contact_FrontGridInterruption}
    \end{subfigure}
    \begin{subfigure}{0.24\linewidth}
        \centering
        \includegraphics[width=\linewidth]{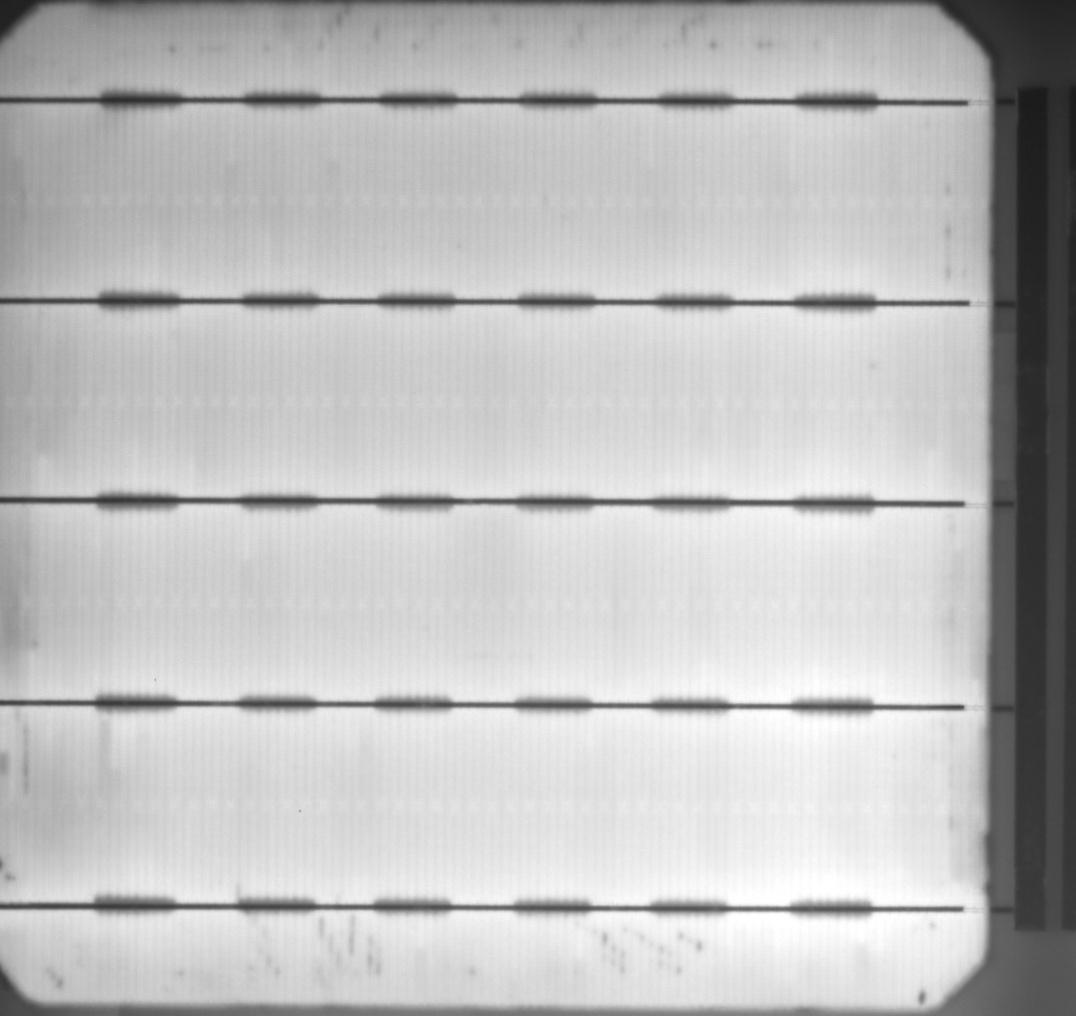}
        \caption[]{\textit{No\_Defect}}
        \label{fig:1d_no_defect}
    \end{subfigure}

    \vspace{3pt}

    \begin{subfigure}{0.24\linewidth}
        \centering
        \includegraphics[width=\linewidth]{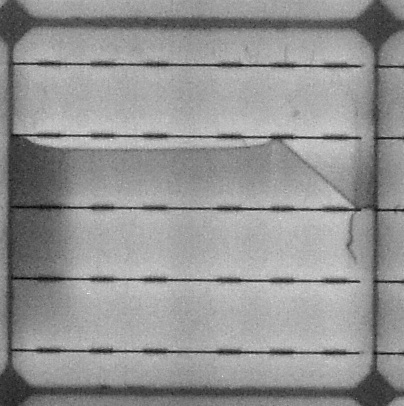}
        \caption[]{Original}
        \label{fig:2a_og}
    \end{subfigure}
    \begin{subfigure}{0.24\linewidth}
        \centering
        \includegraphics[width=\linewidth]{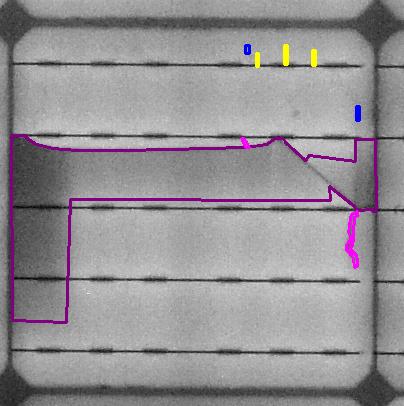}
        \caption[]{Boundaries}
        \label{fig:2b_bound}
    \end{subfigure}
    \begin{subfigure}{0.24\linewidth}
        \centering
        \includegraphics[width=\linewidth]{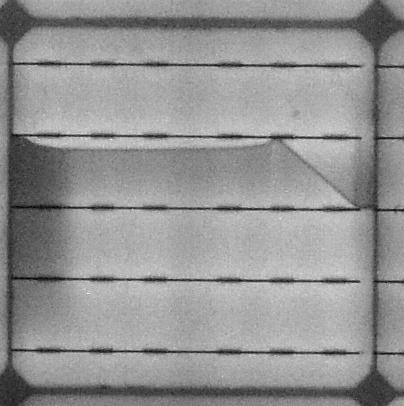}
        \caption[]{\textit{Crack\_Resistive}}
        \label{fig:2c_crack_resistive}
    \end{subfigure}
    \begin{subfigure}{0.24\linewidth}
        \centering
        \includegraphics[width=\linewidth]{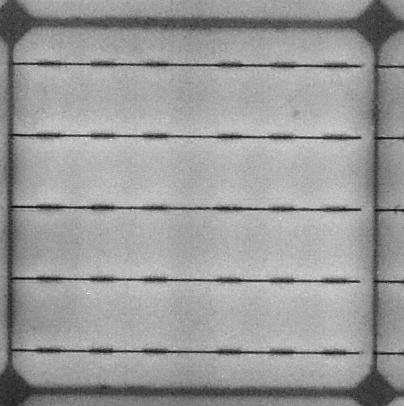}
        \caption[]{\textit{No\_Defect}}
        \label{fig:2d_no_defect}
    \end{subfigure}
    
    \vspace{3pt}

    \begin{subfigure}{0.24\linewidth}
        \centering
        \includegraphics[width=\linewidth]{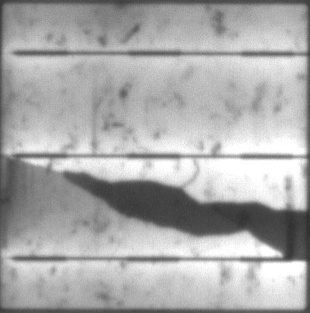}
        \caption[]{Original}
        \label{fig:3a_og}
    \end{subfigure}
    \begin{subfigure}{0.24\linewidth}
        \centering
        \includegraphics[width=\linewidth]{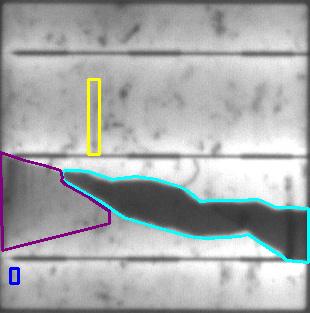}
        \caption[]{Boundaries}
        \label{fig:3b_bound}
    \end{subfigure}
    \begin{subfigure}{0.24\linewidth}
        \centering
        \includegraphics[width=\linewidth]{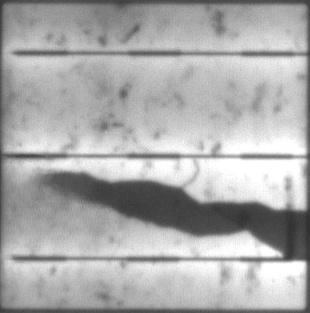}
        \caption[]{\textit{Crack\_Isolated}}
        \label{fig:3c_crack_isolated}
    \end{subfigure}
    \begin{subfigure}{0.24\linewidth}
        \centering
        \includegraphics[width=\linewidth]{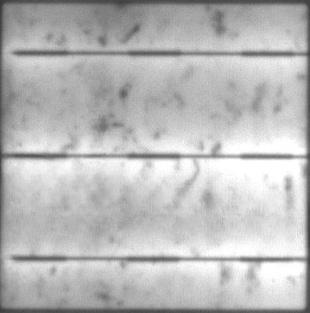}
        \caption[]{\textit{No\_Defect}}
        \label{fig:3d_no_defect}
    \end{subfigure}
    
    \vspace{3pt}

    \begin{subfigure}{0.24\linewidth}
        \centering
        \includegraphics[width=\linewidth]{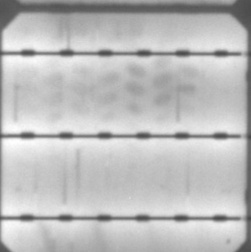}
        \caption[]{Original}
        \label{fig:4a_og}
    \end{subfigure}
    \begin{subfigure}{0.24\linewidth}
        \centering
        \includegraphics[width=\linewidth]{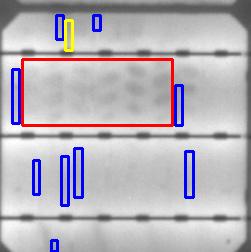}
        \caption[]{Boundaries}
        \label{fig:4b_bound}
    \end{subfigure}
    \begin{subfigure}{0.24\linewidth}
        \centering
        \includegraphics[width=\linewidth]{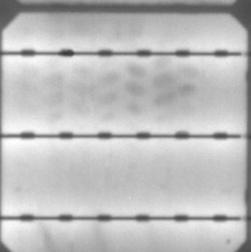}
        \caption[]{\textit{Contact\_BeltMarks}}
        \label{fig:4c_Contact_BeltMarks}
    \end{subfigure}
    \begin{subfigure}{0.24\linewidth}
        \centering
        \includegraphics[width=\linewidth]{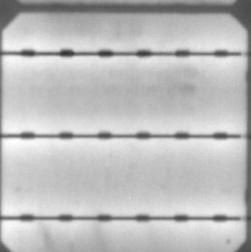}
        \caption[]{\textit{No\_Defect}}
        \label{fig:4d_no_defect}
    \end{subfigure}
    
    \vspace{3pt}

    \caption{\textbf{\textit{Visual examples of the Generative Defect Isolation (GDI) pipeline.}} Each row corresponds to a different source image. For each, we show the original multi-defect cell (left), the annotated defect boundaries and the high-quality samples generated by GDI. These include images with an isolated defect with an area greater than 10\% of the entire image or a clean \textit{No\_Defect} version created by inpainting all defects. Subfigure \ref{fig:1c_Contact_FrontGridInterruption} shows \textit{C\_FGInterruption}, which stands for \textit{Contact\_FrontGridInterruption}. Subfigure \ref{fig:4c_Contact_BeltMarks} displays \textit{Contact\_BeltMarks}, an extremely rare defect class (only 9 original training samples).}

    \label{fig:gdi_samples_fine}
\end{figure}

A qualitative review of the generated samples confirms the high quality and effectiveness of the Generative Defect Isolation (GDI) pipeline, as shown in Figure~\ref{fig:gdi_samples_fine}. The inpainting model excels at producing visually plausible images suitable for training. It demonstrates a strong capability for creating clean \textit{No\_Defect} images by seamlessly removing various defects (Figure~\ref{fig:1d_no_defect}, \ref{fig:2d_no_defect} and \ref{fig:3d_no_defect}). The pipeline is particularly adept at reconstructing key structural features; for example, Figure~\ref{fig:2d_no_defect} illustrates how grid lines that were completely obscured by an original defect are accurately restored, preserving the cell's structural integrity. This ability to generate clean, targeted training data provides a clear advantage and the strong quantitative improvements detailed in the subsequent sections suggest the benefits significantly outweigh potential minor artifacts. A discussion of the method's limitations and failure modes is presented in Section~\ref{sec:discussion}.

\subsection{Experimental Setup}
\label{sec:exp_setup}

\subsubsection{Model Architectures}
To validate our methodology, we selected three powerful and distinct model architectures. We employed two Vision Transformer models, ViT-Small (ViT-S) and ViT-Large (ViT-L), together with the convolutional model EfficientNetV2-L. This selection allows for a comprehensive comparison across different model scales and fundamental architectural paradigms. To leverage transfer learning and accelerate convergence, all models were initialized with weights pretrained on the ImageNet-1k dataset \citep{deng2009imagenet}. This standard practice provides a strong feature-learning foundation for the downstream classification task.

\paragraph{Vision Transformer (ViT)}
The Vision Transformer (ViT) adapts the standard Transformer encoder for image recognition tasks~\citep{dosovitskiy2020image}. An input image is divided into a sequence of flattened, non-overlapping 2D patches, which are linearly projected into a constant latent dimension $D$ to form patch embeddings. Learnable one-dimensional positional embeddings are added to retain spatial information and following the precedent set by BERT~\citep{devlin2019bert}, a learnable \texttt{[class]} token is prepended to the sequence. The final representation of this token is used for image-level classification.

The Transformer encoder processes these embeddings through alternating layers of multi-head self-attention (MSA) and multi-layer perceptron (MLP) blocks, each preceded by Layer Normalization and followed by residual connections. The MLP block comprises two fully connected layers with a GELU non-linearity. Table~\ref{tab:vit_architectures} summarizes the architectural configurations of the ViT models used in this study~\citep{dosovitskiy2020image, steiner2021train}.

\begin{table}[h!]
\centering
\caption{\textbf{\textit{Architectural details for the Vision Transformer (ViT) models used in this study.}}}
\label{tab:vit_architectures}
\resizebox{\textwidth}{!}{%
\begin{tabular}{@{}lccccc@{}}
\toprule
\textbf{Model} & \textbf{Layers} & \textbf{Hidden Size (D)} & \textbf{MLP Size} & \textbf{Attention Heads} & \textbf{Total Parameters} \\ 
\midrule
ViT-S~\cite{steiner2021train} & 12 & 384 & 1536 & 6 & 22.2M \\
ViT-L~\cite{dosovitskiy2020image} & 24 & 1024 & 4096 & 16 & 307M \\ 
\bottomrule
\end{tabular}%
}
\end{table}

\paragraph{EfficientNetV2-L}
EfficientNetV2 is a family of convolutional networks developed using a training-aware neural architecture search (NAS) and compound scaling, designed to jointly optimize accuracy, training speed and parameter efficiency~\citep{tan2021efficientnetv2}. The Large variant (V2-L) contains approximately 120M parameters and incorporates several key architectural refinements.

\begin{itemize}
    \item \textbf{Fused-MBConv Blocks:} The network employs both standard MBConv (Mobile Inverted Bottleneck Convolution) and Fused-MBConv blocks, with the latter used predominantly in the early stages. An MBConv block expands channel dimensions via a $1\times1$ convolution, applies a depthwise $3\times3$ convolution and then projects back to a lower-dimensional space, forming an efficient inverted bottleneck structure. The Fused-MBConv variant merges the depthwise and expansion convolutions into a single $3\times3$ convolution, improving computational efficiency on modern hardware accelerators.
    \item \textbf{Architectural Refinements:} The NAS-derived design favors smaller expansion ratios within MBConv blocks to reduce memory access overhead and uses smaller $3\times3$ kernels, compensating for the reduced receptive field by increasing network depth. The final stage consists of a $1\times1$ convolution, global pooling and a fully connected layer for classification.
    \item \textbf{Scaling Strategy:} To obtain larger variants such as EfficientNetV2-L, a non-uniform compound scaling strategy is applied, progressively adding layers to later stages of the network to increase representational capacity without incurring significant runtime overhead.
\end{itemize}

These design choices collectively enable EfficientNetV2-L to achieve strong performance while maintaining high computational efficiency~\citep{tan2021efficientnetv2}.

\subsubsection{Data Splits and Evaluation}

To simulate data-scarce conditions, we created subsets of the training data containing 1\%, 5\%, 10\%, 20\% and 50\% of the full training set. Inpainted samples within each subset were generated only from images already included in that subset. Training was performed without any threshold tuning. At inference time, each class probability was converted to a binary prediction using a per-class threshold; for the main experiments, these thresholds were selected per class as the value that maximized the macro F1-score on the test set. The \emph{identical} procedure was applied to the baseline and GDI models to preserve a fair comparison. Since this selection used the test set, it can yield optimistic absolute F1 estimates, although it does not affect model training. To show that GDI improvements are threshold independent, Section~\ref{sec:robustness} reports a complementary evaluation at a fixed threshold ($\tau = 0.5$) on the same threshold-free models, confirming the gains are a property of the augmentation rather than of threshold tuning. All reported results are computed using the two multi-label metrics defined below.

\begin{itemize}
    \item \textbf{Zero-One Accuracy:} This strict metric measures the fraction of samples for which the set of predicted labels is exactly identical to the set of true labels. It is calculated as:
    $$ \text{Accuracy}_{0/1} = \frac{1}{N} \sum_{i=1}^{N} \mathbb{I}(\hat{y}_i = y_i) $$
    where $N$ is the number of samples, $y_i$ is the true label set, $\hat{y}_i$ is the predicted label set and $\mathbb{I}(\cdot)$ is the indicator function.

    \item \textbf{Macro F1-Score:} This metric provides a balanced measure of precision and recall, averaged across all classes, making it ideal for imbalanced datasets. For each class $c$ in the set of all classes $C$, the F1-Score is the harmonic mean of precision ($P_c$) and recall ($R_c$). The macro-averaged F1-Score is their unweighted average:
    $$ P_c = \frac{\text{TP}_c}{\text{TP}_c + \text{FP}_c} \quad \quad R_c = \frac{\text{TP}_c}{\text{TP}_c + \text{FN}_c} $$
    $$ F1_c = 2 \cdot \frac{P_c \cdot R_c}{P_c + R_c} $$
    $$ \text{Macro F1} = \frac{1}{|C|} \sum_{c \in C} F1_c $$
\end{itemize}

\subsubsection{Data Augmentation and Implementation Details}

To ensure robust feature learning and a fair comparison between methods, an identical data augmentation pipeline was applied to all models (Baseline and GDI) during training. To standardize the highly variable native resolutions described in Section~\ref{sec:methodology_dataset_and_init_proc}, all images across both the training and test sets were resized to a uniform $224 \times 224$ pixels prior to model input. The training transformations included: random resized crop to $224 \times 224$ pixels, random horizontal flip, color jitter and random affine transformations. Input images were normalized using the standard ImageNet mean and standard deviation values.

To ensure that any observed performance gains were solely due to the improved data quality, we maintained strict consistency in our training configuration. For each model architecture, identical hyperparameters, including learning rate, batch size, optimizer choice and weight decay were used for both the baseline and the GDI augmented training runs.

\subsection{Overall Performance Gains}
The results clearly and consistently demonstrate that training with the GDI-generated dataset provides a significant performance advantage over the baseline across nearly all experimental configurations. The GDI models show improved accuracy and F1-Scores, validating our hypothesis that isolating defects helps resolve learning ambiguities and enhances feature disentanglement.

\begin{figure}[htb!]
    \centering
    \begin{subfigure}{0.33\textwidth}
        \centering
        \includegraphics[width=\linewidth]{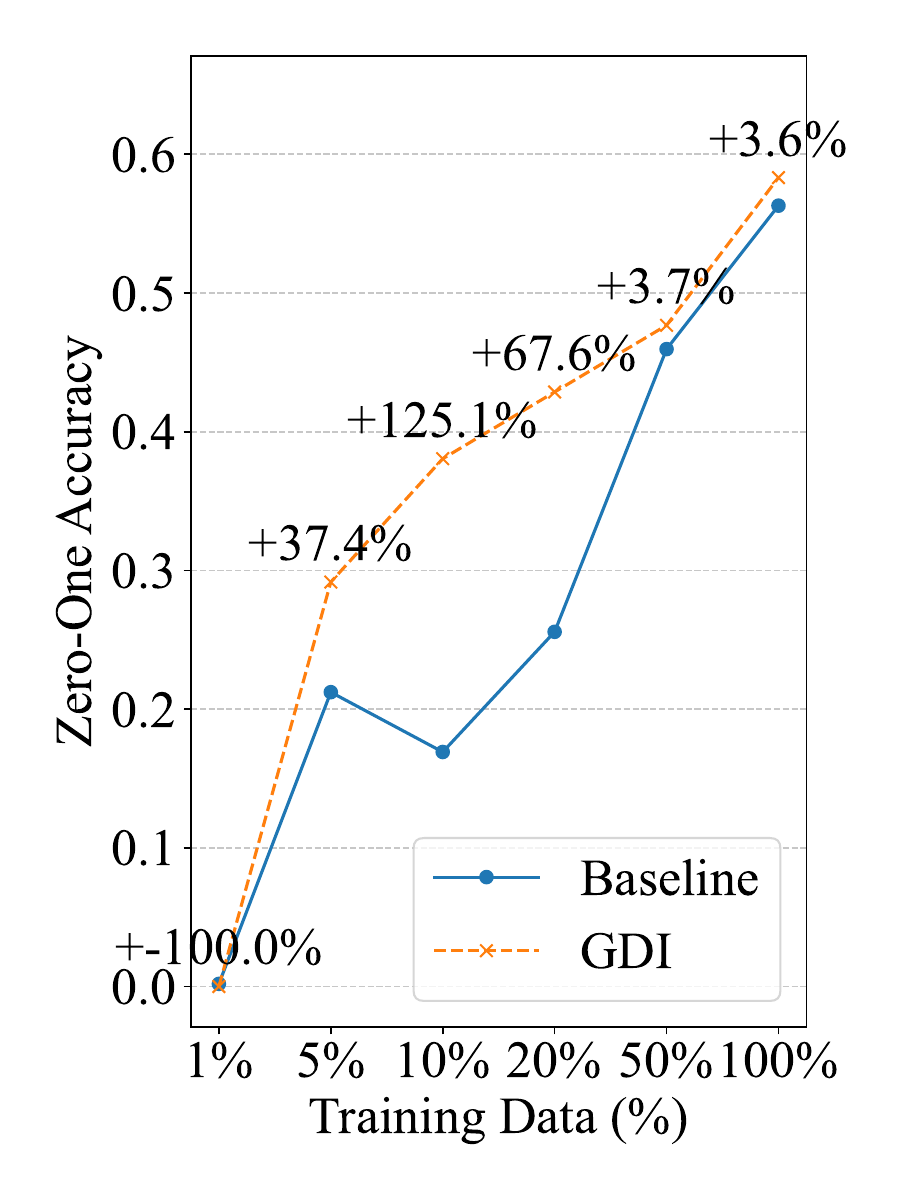}
        \includegraphics[width=\linewidth]{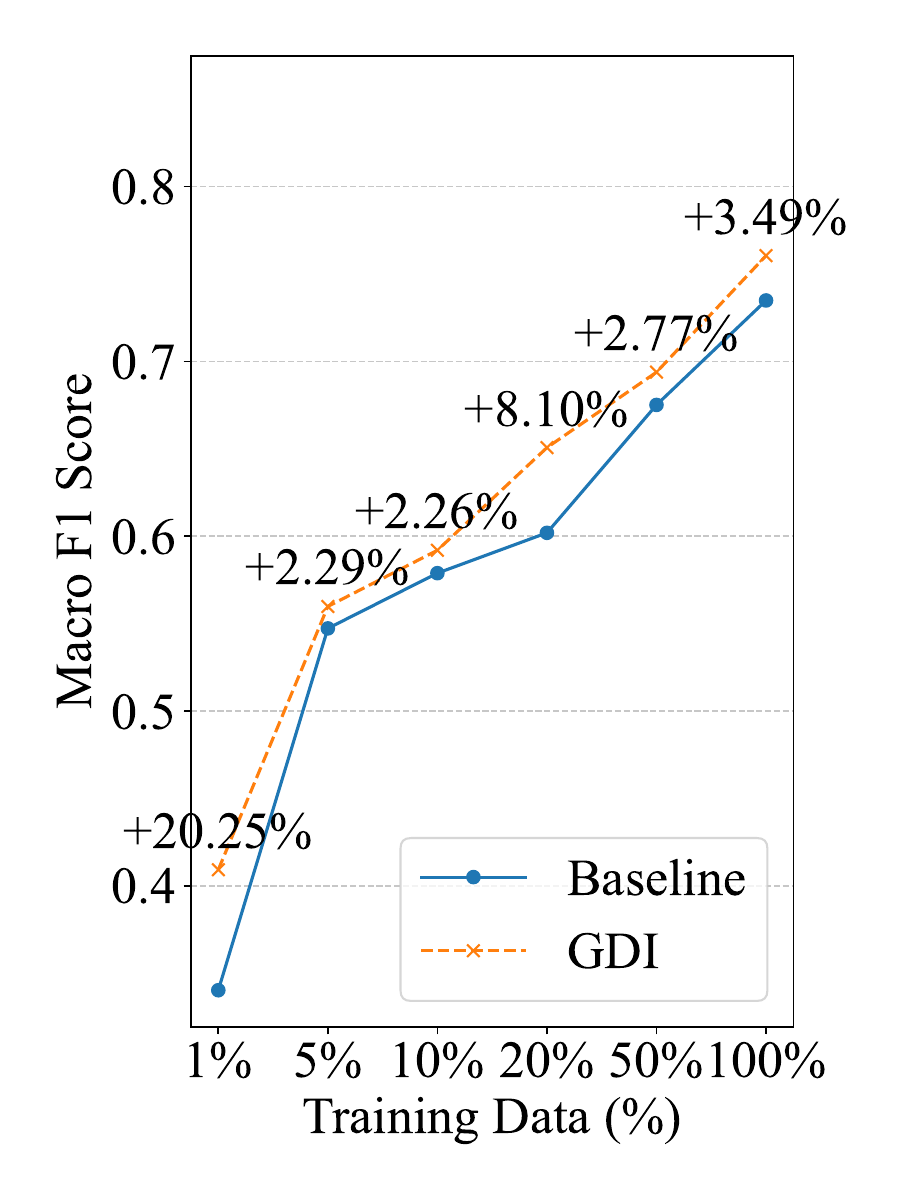}
        \caption{ViT-S}
    \end{subfigure}\hfill
    \begin{subfigure}{0.33\textwidth}
        \centering
        \includegraphics[width=\linewidth]{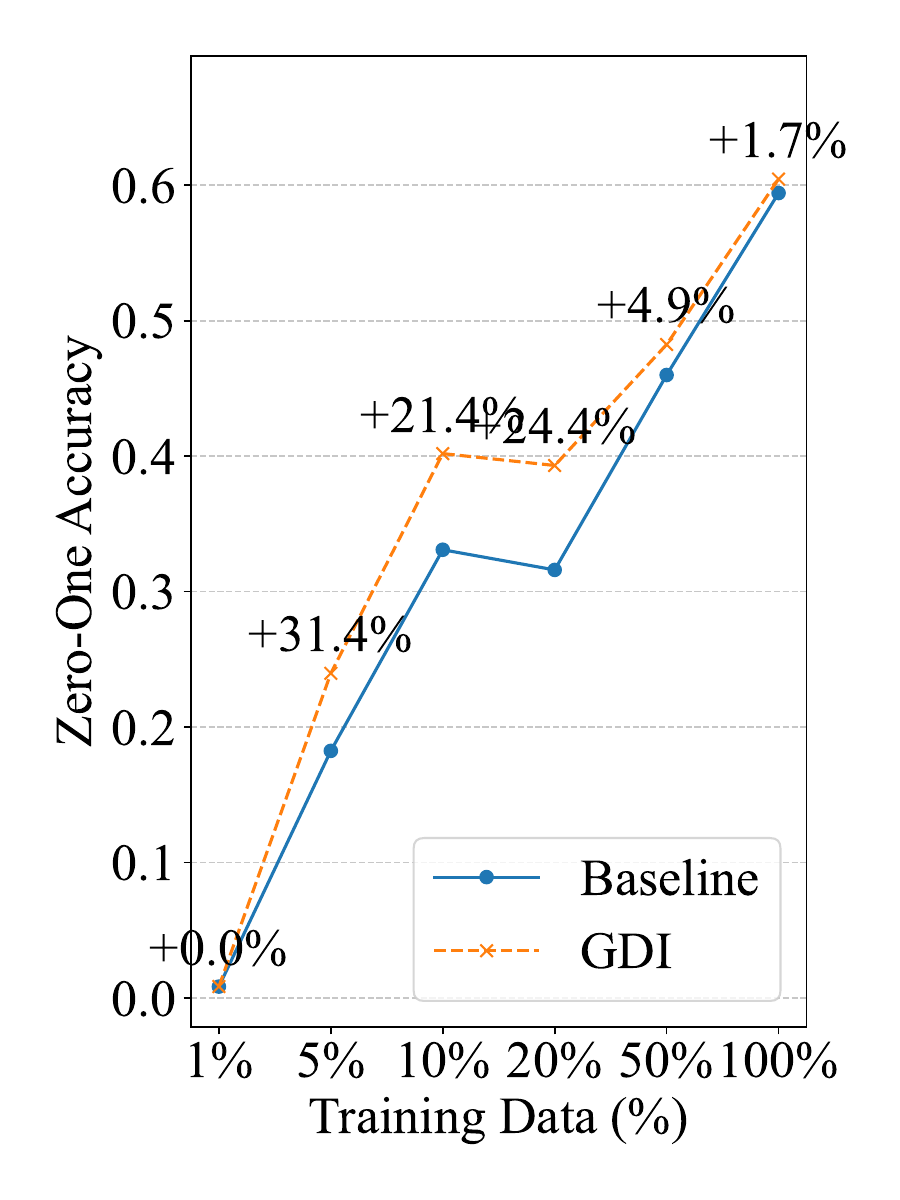}
        \includegraphics[width=\linewidth]{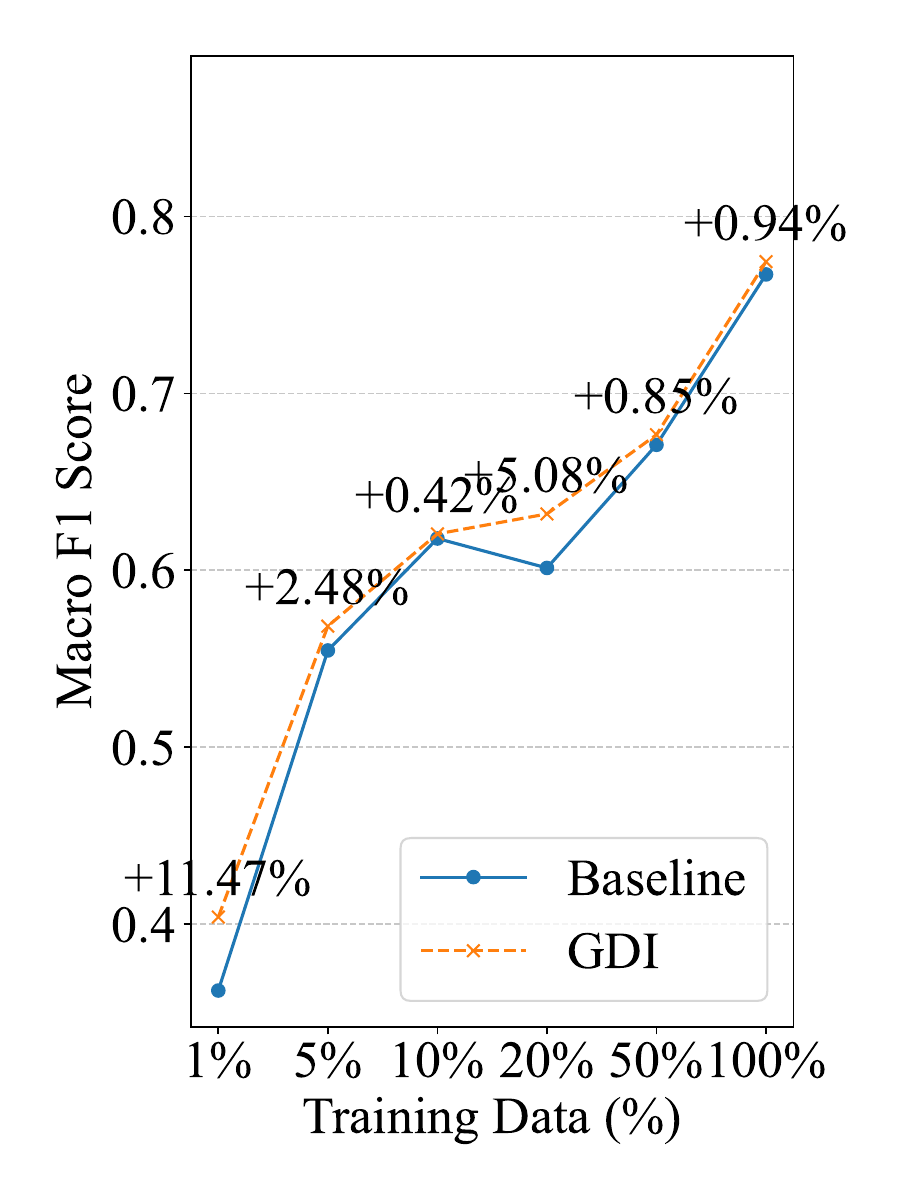}
        \caption{EfficientNetV2-L}
    \end{subfigure}\hfill
    \begin{subfigure}{0.33\textwidth}
        \centering
        \includegraphics[width=\linewidth]{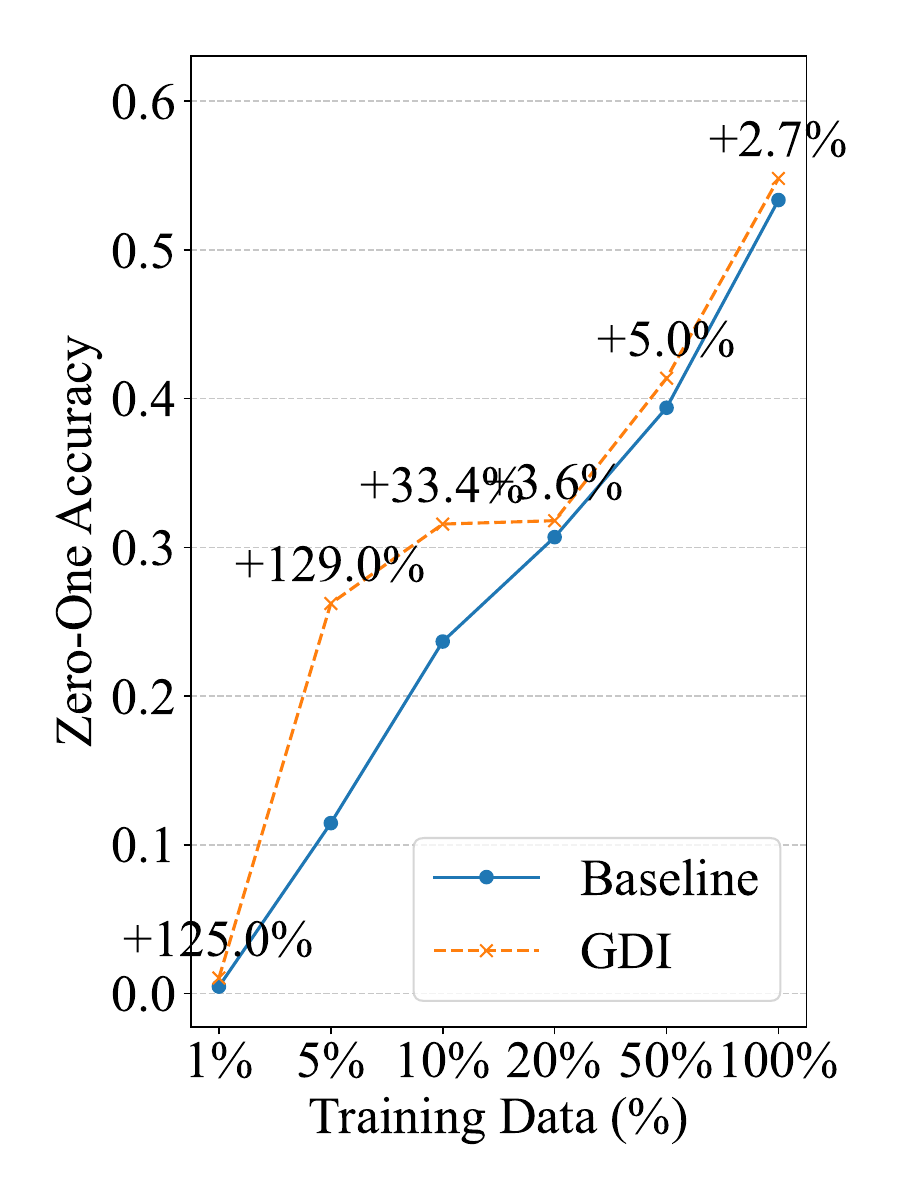}
        \includegraphics[width=\linewidth]{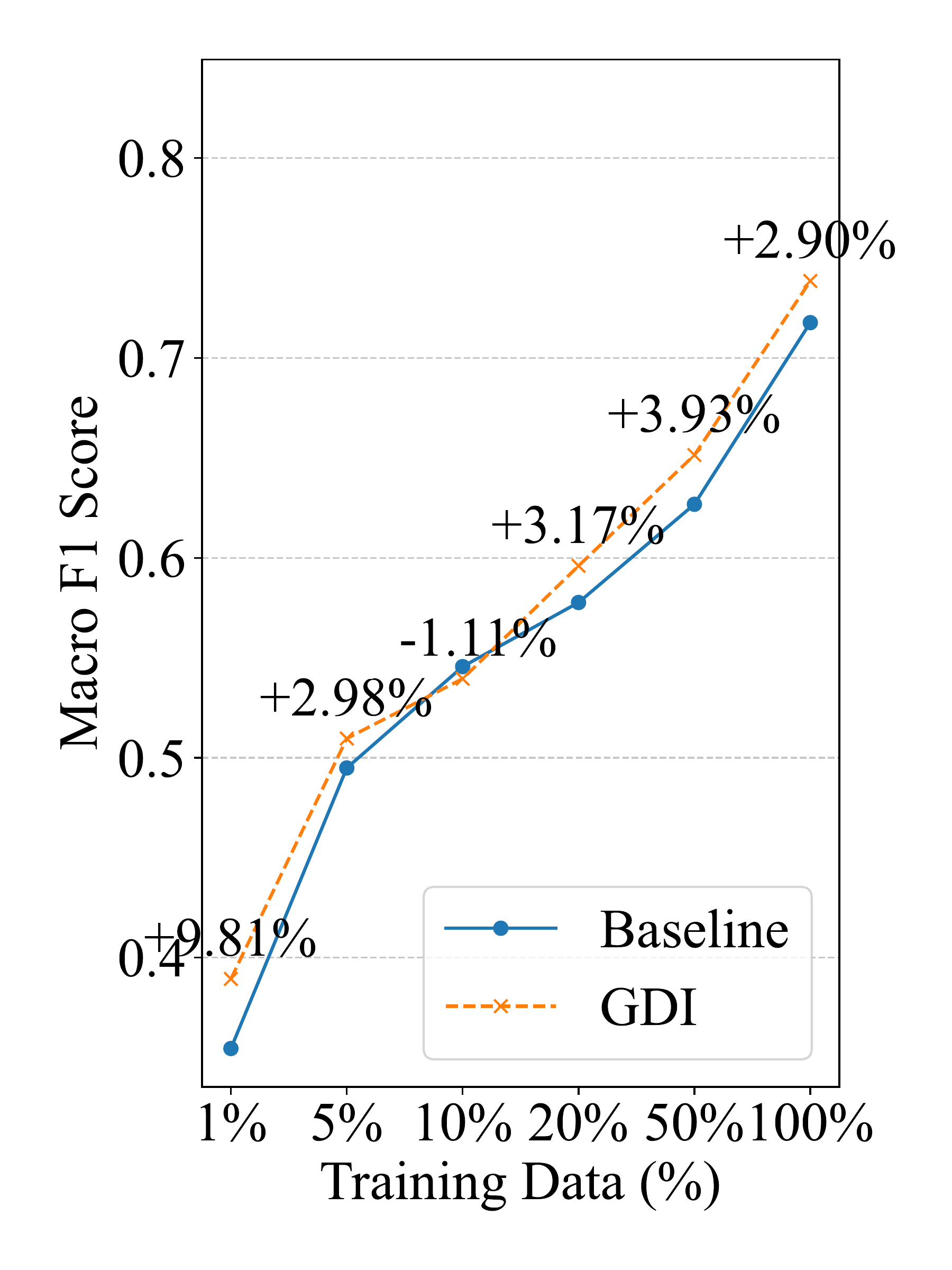}
        \caption{ViT-L}
    \end{subfigure}
    \caption{\textbf{\textit{Performance comparison of the baseline and GDI models across three architectures for the six different training splits: 1\%, 5\%, 10\%, 20\%, 50\%, 100\%}}. The top image in each column shows Zero-One Accuracy and the bottom image shows Macro F1 Score.}
    \label{fig:model_comparisons}
\end{figure}

As shown in Figure~\ref{fig:model_comparisons}, the GDI-augmented models almost universally outperform their baseline counterparts in Zero-One Accuracy. A notable outlier is the 100\% drop in accuracy for the GDI-augmented ViT-S model at the 1\% data level. This reflects the accuracy falling to zero, but it is crucial to note the baseline was exceedingly small (0.0019), rendering the relative metric misleading in this specific, data-scarce instance. Importantly, this exception aside, the GDI-augmented models also consistently achieve higher F1-Scores than their baseline counterparts (Figure~\ref{fig:model_comparisons}).

\subsection{Impact of Data Scarcity}
A key finding is that the benefits of GDI are most pronounced in low-data settings. When the training set is limited, providing the model with clean, unambiguous, single-defect examples proves to be exceptionally effective.

For \textbf{Zero-One Accuracy}, the improvements in data-scarce scenarios are substantial. For instance, with only 10\% of the training data, the ViT-S model sees a remarkable \textbf{+125.1\%} relative improvement. Similarly, the ViT-L model achieves massive gains of \textbf{+129.0\%} with just 5\% of the data. This strongly suggests that GDI is a powerful tool for boosting performance when labeled data is expensive or difficult to acquire.

This trend is mirrored in the \textbf{F1-Scores}. At the 1\% data level, ViT-S shows a substantial \textbf{+20.3\%} improvement, while EfficientNetV2-L and ViT-L also see significant gains of \textbf{+11.5\%} and \textbf{+9.8\%}, respectively. As the amount of training data increases, the relative gains naturally become more modest, yet they remain consistently positive.

\subsection{Performance Across Model Architectures}

Our proposed GDI method shows robust improvements across all tested model architectures, from the lightweight ViT-S to the large-scale ViT-L and the convolutional EfficientNetV2-L.

Interestingly, the smaller \textbf{ViT-S} model often reaps the largest \textit{relative} benefits from GDI, especially in accuracy (e.g., +125.1\% at 10\% data) and F1-Score (+20.3\% at 1\% data). This suggests that GDI is particularly effective for models with fewer parameters, which may have less intrinsic capacity to disentangle complex, overlapping features. In contrast, the powerful \textbf{EfficientNetV2-L} consistently achieves the highest absolute scores in higher-data settings (50\% and 100\%), establishing the best overall performance with a top F1-Score of \textbf{0.7744}. For this model, the improvements from GDI are more incremental but consistently positive, demonstrating that even a highly capable architecture benefits from cleaner training signals. The large \textbf{ViT-L} model follows a similar positive trend, though we noted one minor instance of F1-Score degradation (-1.1\% at 10\% data), which we attribute to potential training instability for a very large model on a limited data subset.

A key trend emerging from these results is that our inpainting-based augmentation exhibits diminishing utility as the volume of training data increases. In a low-data setting, the augmentation provides a significant performance advantage by synthetically expanding under-represented classes and providing idealized, representative examples. This simplifies the feature-learning task by removing confounding visual information, which is critical when real examples are sparse. As the dataset size approaches 100\%, however, the performance of the augmented and baseline models converges. With access to the complete dataset, the baseline model can learn to distinguish complex features from real multi-defect images organically. Consequently, the marginal benefit of the synthetic data is reduced, as the richness of the full real dataset becomes more impactful for generalization.


\subsection{Class-Wise Performance Analysis}

While aggregate metrics confirm the overall benefit of GDI, a deeper, class-level analysis reveals a more nuanced picture of its impact. Although the ViT-S model achieved the highest Zero-One Accuracy (0.4286, a 67.6\% improvement over its baseline) and Macro F1-Score (0.6507, up 8.10\% from baseline) on the 20\% data split, we focus our analysis on the \textbf{EfficientNetV2-L} model. The 20\% split was chosen as it represents a realistic, low-data scenario, allowing us to better assess the impact of GDI. EfficientNetV2-L was selected because it not only achieved the strongest overall performance on the full dataset, providing a reliable baseline, but also demonstrated substantial gains in this data-constrained setting (\textbf{+24.4\%} in Zero-One Accuracy and \textbf{+5.1\%} in Macro F1-Score), making it both powerful and representative for class-level analysis.

\begin{figure}[htbp]
    \centering
    \includegraphics[width=0.98\textwidth]{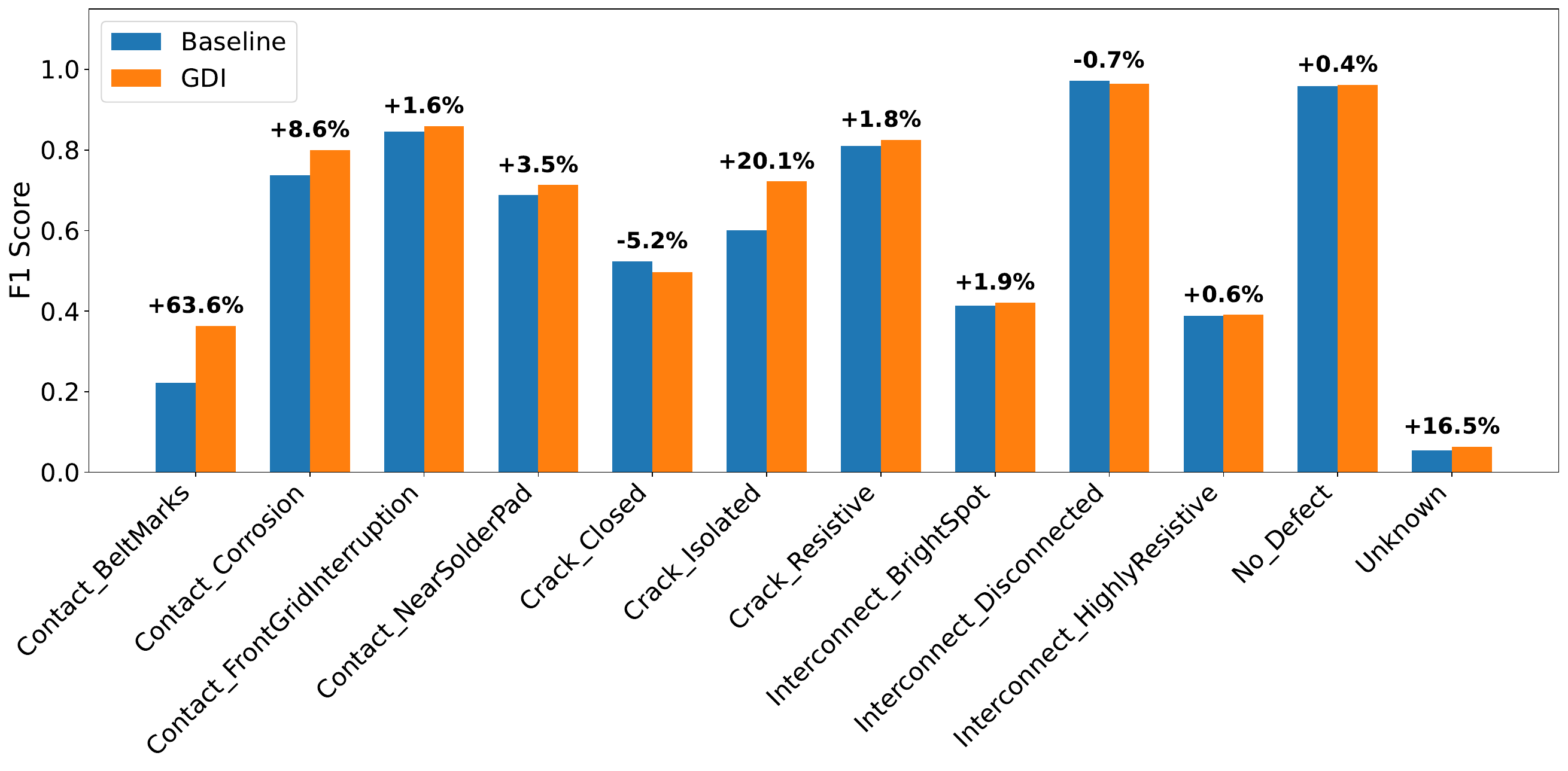}
    \caption{\textbf{\textit{Class-wise F1-Scores:}} Comparison between the baseline and GDI-enhanced EfficientNetV2-L models on the 20\% data split, showing gains for rare defects like \textit{Contact\_BeltMarks} and \textit{Contact\_Corrosion}.}
    \label{fig:class_wise_f1_score_comparison}
\end{figure}

The composition of the GDI training set and its performance impact for this experiment is detailed in Table \ref{tab:combined_20}, while the corresponding comparison of baseline and GDI F1-Scores is illustrated in Figure \ref{fig:class_wise_f1_score_comparison}. The GDI model demonstrates substantial improvements, particularly for classes that were either rare or frequently co-occurred with other defects in the original dataset.

The most significant improvement is seen in \textit{Contact\_BeltMarks} (visualized in Figure \ref{fig:4c_Contact_BeltMarks}), which saw its F1-Score increase by \textbf{+63.6\%}. This class was extremely rare in the original 20\% split and the addition of just 3 clean, isolated examples from GDI was enough to provide a more stable learning signal.

Significant gains are also observed for other visually ambiguous or less common classes. The performance for \textit{Crack\_Isolated} improved by \textbf{+20.1\%}, a direct result of its training set being bolstered with clean, isolated examples from GDI. Interestingly, the F1-Score for the \textit{Unknown} class also saw a notable \textbf{+16.5\%}. This improvement is an indirect benefit of GDI; although no new samples were generated for \textit{Unknown}, by training on unambiguous examples of other defects, the model becomes more decisive in recognizing them. This reduces its tendency to misclassify known defects as \textit{Unknown}, leading to a more precise application of the label.

For classes that were already well-represented and had high baseline scores, GDI provided more modest gains. For instance, \textit{No\_Defect} and \textit{Contact\_FrontGridInterruption} already achieved high baseline F1-Scores of 0.9586 and 0.8458, respectively, indicating the model had a strong existing understanding of these classes. For such well-learned categories, the marginal benefit of adding more clean examples is naturally lower compared to rare or ambiguous classes where the baseline model initially struggled.

We note a minor decrease in F1-Score for two classes: \textit{Crack\_Closed} (-5.2\%) and \textit{Interconnect\_Disconnected} (-0.7\%). The latter's score was already near-perfect in the baseline (0.9718), making further improvement difficult. The drop for \textit{Crack\_Closed} might be attributed to the relatively small number of GDI samples added (2.6\%) being insufficient to overcome the variance in this specific training run. However, these isolated instances are exceptions to the overwhelmingly positive trend. The results strongly indicate that GDI is a highly effective strategy for boosting the performance on rare and hard-to-distinguish classes by providing clean, unambiguous training signals.

\begin{table}[h!]
\centering
\caption{\textbf{\textit{Training data composition and relative F1-Score change for the 20\% data experiment.}} The table shows the number of original multi-defect labels, the number of single-defect inpainted samples generated by GDI for each class and the relative F1-Score change compared to the baseline.}
\label{tab:combined_20}
\resizebox{\textwidth}{!}{%
\begin{tabular}{lrrrrr}
\toprule
\textbf{Defect Class} & \makecell{\textbf{Original}\\\textbf{Labels}} & \makecell{\textbf{Inpainted}\\\textbf{Samples}} & \makecell{\textbf{Total}\\\textbf{Samples}} & \makecell{\textbf{Inpainted}\\\textbf{\%}} & \makecell{\textbf{Relative}\\\textbf{F1 Change (\%)}} \\ 
\midrule
Contact\_BeltMarks & 6 & 3 & 9 & 33.3\% & \textbf{+63.6\%} \\
Contact\_Corrosion & 19 & 7 & 26 & 26.9\% & \textbf{+8.6\%} \\
Contact\_FrontGridInterruption & 1953 & 13 & 1966 & 0.7\% & \textbf{+1.6\%} \\
Contact\_NearSolderPad & 1229 & 15 & 1244 & 1.2\% & \textbf{+3.5\%} \\
Crack\_Closed & 442 & 12 & 454 & 2.6\% & -5.2\% \\
Crack\_Isolated & 288 & 58 & 346 & 16.8\% & \textbf{+20.1\%} \\
Crack\_Resistive & 595 & 434 & 1029 & 42.2\% & \textbf{+1.8\%} \\
Interconnect\_BrightSpot & 114 & 56 & 170 & 32.9\% & \textbf{+1.9\%} \\
Interconnect\_Disconnected & 32 & 5 & 37 & 13.5\% & -0.7\% \\
Interconnect\_HighlyResistive & 114 & 12 & 126 & 9.5\% & \textbf{+0.6\%} \\
No\_Defect & 119 & 1633 & 1752 & 93.2\% & \textbf{+0.4\%} \\
Unknown & 14 & 0 & 14 & 0.0\% & \textbf{+16.5\%} \\
\midrule
\textbf{Total / Average} & \textbf{4925} & \textbf{2248} & \textbf{7173} & \textbf{22.7\%} & \textbf{+9.4\%} \\
\bottomrule
\end{tabular}%
}
\end{table}

\begin{figure}[!htb]
    \centering
    \begin{subfigure}[b]{0.489\textwidth}
        \centering
        \includegraphics[width=\textwidth]{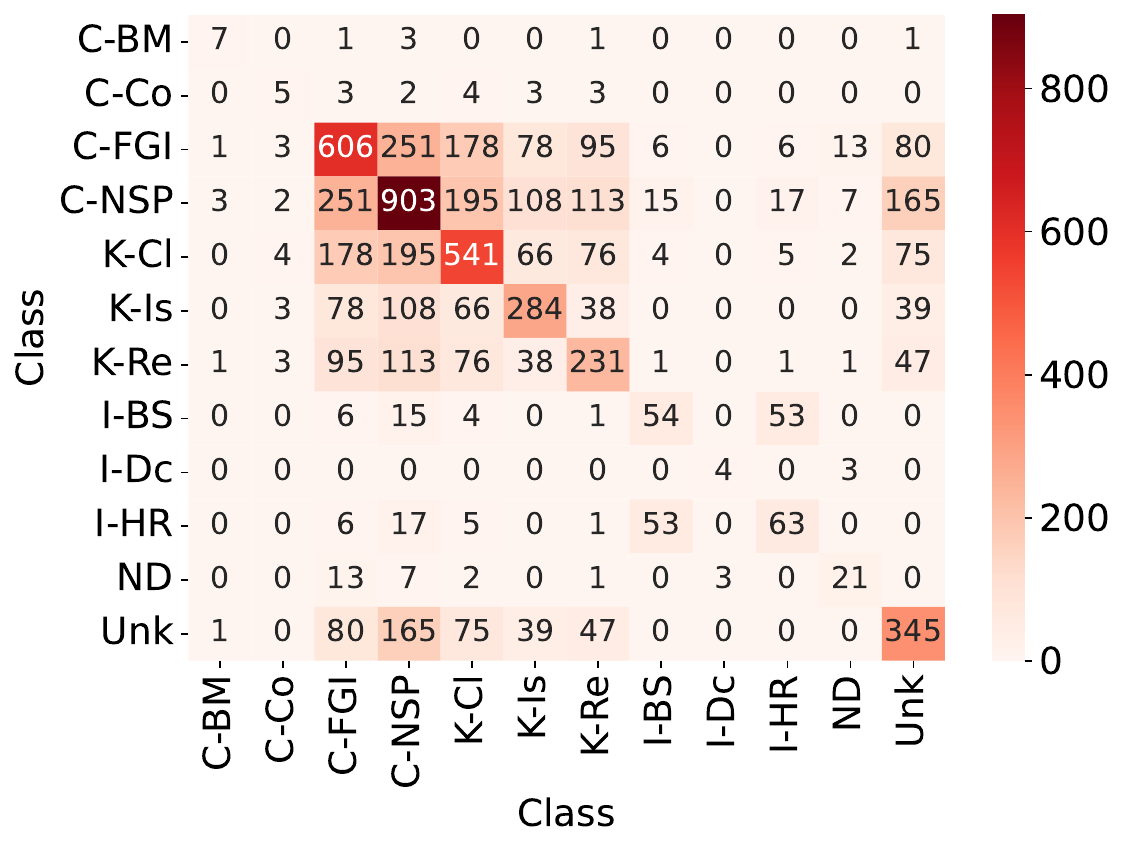}
        \caption{Baseline Model}
        \label{fig:baseline_co_occurrence}
    \end{subfigure}
    \hfill
    \begin{subfigure}[b]{0.489\textwidth}
        \centering
        \includegraphics[width=\textwidth]{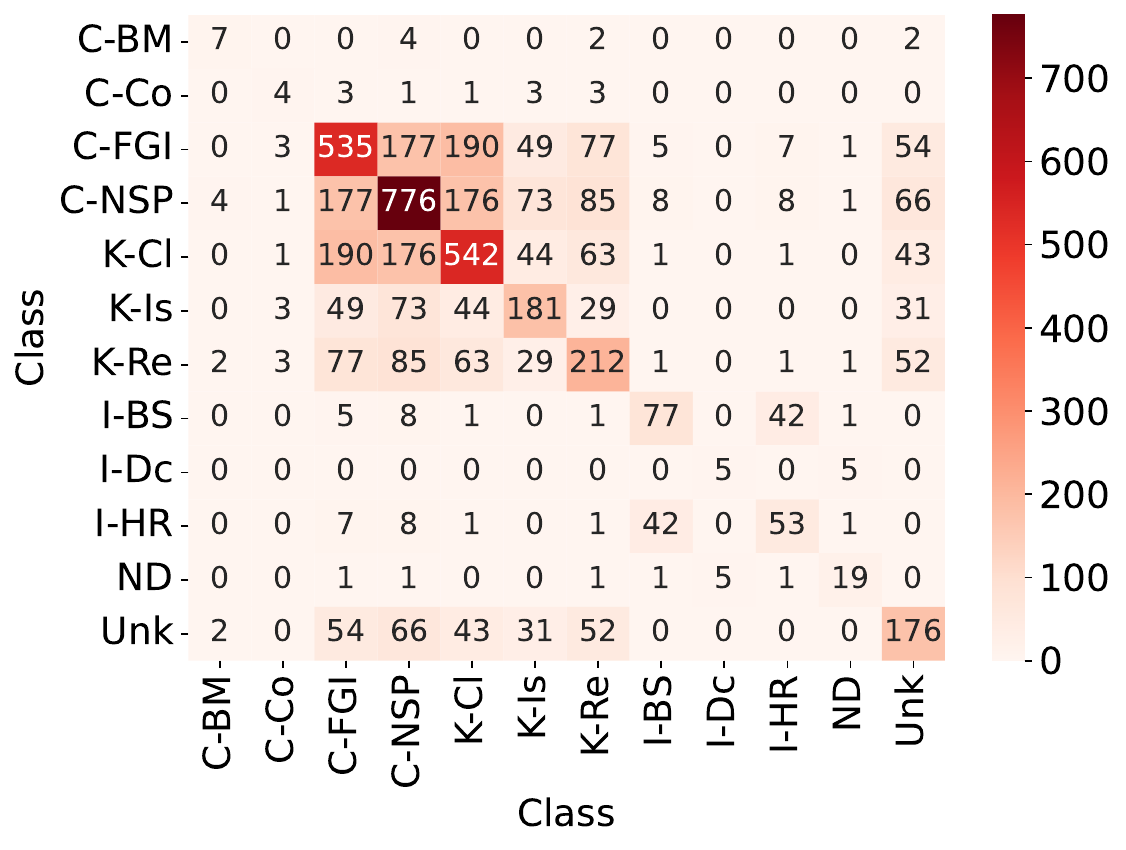}
        \caption{GDI-augmented Model}
        \label{fig:dia_co_occurrence}
    \end{subfigure}
    \caption{\textbf{\textit{Error co-occurrence matrices for the EfficientNetV2-L model on the 20\% data split.}} Each cell $(i, j)$ shows the number of times class $i$ and class $j$ were both incorrectly predicted for the same image. 
    The GDI-augmented model (b) shows a marked reduction in error counts across many class pairs compared to the baseline (a), indicating less confusion. 
    Abbreviations: C-BM (Contact\_BeltMarks), C-Co (Contact\_Corrosion), C-FGI (Contact\_FrontGridInterruption), C-NSP (Contact\_NearSolderPad), 
    K-Cl (Crack\_Closed), K-Is (Crack\_Isolated), K-Re (Crack\_Resistive), 
    I-BS (Interconnect\_BrightSpot), I-Dc (Interconnect\_Disconnected), I-HR (Interconnect\_HighlyResistive), 
    ND (No\_Defect), Unk (Unknown).}
    \label{fig:error_co_occurrence}
\end{figure}

To further investigate how GDI helps the model disentangle features, we analyzed the co-occurrence of errors. We generated a matrix for each model where both axes represent the defect classes (Figure~\ref{fig:error_co_occurrence}). Each off-diagonal cell $(i, j)$ in the matrix counts the number of samples where the model made an error on Class~$i$ and \textit{simultaneously} made an error on Class~$j$. The diagonal cells simply show the total number of errors for that specific class. A lower off-diagonal count in the GDI matrix therefore signifies reduced confusion between that pair of classes.

The comparison reveals that the GDI-augmented model is significantly less confused. The most compelling evidence is the reduced co-occurrence of errors with the \textit{Unknown} class. In the baseline model, when an image contained a \textit{Contact\_NearSolderPad} defect, the model would co-predict \textit{Unknown} 165 times. The GDI model reduces this specific confusion to just 66 instances - a 60\% decrease. This demonstrates that by learning from cleaner, isolated examples, the model becomes more precise in its predictions and less reliant on the ambiguous \textit{Unknown} label. Similarly, confusion between \textit{Unknown} and \textit{Contact\_FrontGridInterruption} dropped from 80 to 54 cases and with \textit{Crack\_Closed}, it fell from 75 to 43 cases.

Beyond the \textit{Unknown} class, GDI also mitigates confusion between distinct defect types. For example, the baseline model frequently confused \textit{Contact\_FrontGridInterruption} and \textit{Contact\_NearSolderPad}, with 251 co-occurring errors. The GDI model reduces this number to 177, a drop of nearly 30\%.

This targeted improvement is reflected across the board: the total number of co-occurring error pairs (the sum of the upper triangle of the matrix) fell from 1,774 in the baseline to 1,312 with GDI, a systematic reduction of 26\%. Collectively, this visual and quantitative evidence strongly supports our central hypothesis: by removing confounding signals from multi-defect images, GDI enables the model to learn more discriminative features, leading to more precise and reliable classifications.

\subsection{Comparison with Other Methods}
To contextualize the performance of our GDI-enhanced models, we compare them against several recent methods that have utilized the same dataset. Table~\ref{tab:results} presents this comparison, evaluating performance on both a 20\% data-scarce split and the full 100\% training dataset. Crucially, we extended our experimentation to apply the GDI augmentation pipeline to these baseline architectures as well, isolating the impact of the data augmentation from the model architecture.

The methodologies from other works were adapted for a fair comparison within our multi-label classification framework. Fioresi et al.~\cite{fioresi2022automated} originally focused on semantic segmentation using DeepLabv3 \citep{chen2017rethinking}; for our purposes, we evaluate the performance of its ResNet-50 \citep{he2016deep} backbone on the classification task. Abdelsattar et al.~\cite{abdelsattar2025automated_baselines} benchmarked multiple models for binary classification of defects and we include their two best-performing architectures, MobileNetV2 \citep{sandler2018mobilenetv2} and Xception \citep{chollet2017xception}, in our comparison.

The results in Table~\ref{tab:results} demonstrate that the benefits of GDI are broadly model-agnostic. In the 20\% data-scarce regime, every tested architecture from the lightweight MobileNetV2 to the heavier ResNet-50 achieved higher Accuracy and F1-Scores when trained with GDI compared to their standard baselines. For instance, the ResNet-50 backbone from \cite{fioresi2022automated} saw its F1-Score rise from 0.6119 to 0.6313, while Xception improved from 0.6073 to 0.6195. This universality confirms that the synthetic isolation of defects provides a cleaner training signal that aids feature disentanglement regardless of the underlying inductive bias of the network. At the 100\% data scale, GDI continues to provide marginal to notable gains for most architectures, with Xception seeing its F1-Score jump to 0.7569 from 0.7305.

Ultimately, our proposed architectures equipped with GDI establish a new state-of-the-art. In the low-data scenario (20\%), our GDI-enhanced ViT-S model achieves the highest accuracy (0.4286) and F1-Score (0.6507), significantly outperforming all prior methods even when those methods are also augmented with GDI. On the full dataset, our EfficientNetV2-L (GDI) secures the top overall performance with an accuracy of 0.6046 and an F1-Score of 0.7744. These results underscore that while GDI acts as a general performance booster, its combination with modern architectures like Vision Transformers and EfficientNetV2-L yields the most robust solution for multi-label defect classification.

\begin{table*}[t]
\centering
\caption{\textbf{\textit{Benchmarking against existing works.}}
Performance comparison of the architectures explored in this study—\textbf{EfficientNetV2-L, ViT-S and ViT-L}—against established baselines. Baseline models are compared with their GDI-augmented counterparts. We evaluate \textbf{ResNet-50} (backbone adopted from \cite{fioresi2022automated}) and the best-performing models from \cite{abdelsattar2025automated_baselines} (\textbf{MobileNetV2 and Xception}). All models are evaluated on the original and GDI-augmented datasets using 20\% and 100\% training splits. Performance is measured by Zero-One Accuracy and Macro F1-score, with best results shown in bold.}
\label{tab:results}

\resizebox{\textwidth}{!}{%
\begin{tabular}{l c c c c}
\hline
\textbf{Model} & \multicolumn{2}{c}{\textbf{20\% Data}} & \multicolumn{2}{c}{\textbf{100\% Data}} \\
 & Accuracy & F1-Score & Accuracy & F1-Score \\
\hline

\multicolumn{5}{l}{\textbf{Baselines}} \\
\hline
ResNet-50 (backbone adopted from \cite{fioresi2022automated})
 & 0.3210 & 0.6119 & 0.5729 & 0.7578 \\
MobileNetV2 \cite{abdelsattar2025automated_baselines}
 & 0.2340 & 0.5866 & 0.5607 & 0.7469 \\
Xception \cite{abdelsattar2025automated_baselines}
 & 0.3111 & 0.6073 & 0.5576 & 0.7305 \\
EfficientNetV2-L
 & 0.3160 & 0.6013 & 0.5943 & 0.7672 \\
ViT-S
 & 0.2557 & 0.6019 & 0.5630 & 0.7349 \\
ViT-L
 & 0.3069 & 0.5777 & 0.5336 & 0.7176 \\
\hline

\multicolumn{5}{l}{\textbf{GDI-augmented (Ours)}} \\
\hline
ResNet-50
 & 0.3687 & 0.6313 & 0.5878 & 0.7584 \\
MobileNetV2
 & 0.2603 & 0.5921 & 0.5638 & 0.7504 \\
Xception
 & 0.3195 & 0.6195 & 0.5676 & 0.7569 \\
EfficientNetV2-L
 & 0.3931 & 0.6318 & \textbf{0.6046} & \textbf{0.7744} \\
ViT-S
 & \textbf{0.4286} & \textbf{0.6507} & 0.5832 & 0.7605 \\
ViT-L
 & 0.3179 & 0.5960 & 0.5481 & 0.7385 \\
\hline
\end{tabular}%
}
\end{table*}

\subsection{Comparison with Copy-Paste Augmentation}

To rigorously validate the effectiveness of our proposed method, we compared Generative Defect Isolation (GDI) against Copy-Paste augmentation \citep{ghiasi2021simple}, a widely adopted technique in instance segmentation. To ensure a fair and robust baseline, we aligned the augmented dataset size exactly with GDI and utilized ground-truth segmentation masks to extract precise defect regions, avoiding background noise contamination. The pasting mechanism was randomized across the training distribution to maximize variance. Crucially, the ground-truth labels were dynamically updated to reflect the union of the source and target defects (e.g., a `Crack' pasted onto a    `Corrosion' sample results in a multi-label entry containing both classes).

As shown in Table~\ref{tab:augmentation_results}, our GDI approach demonstrates superior stability, consistently improving Accuracy and F1-Scores across all three model architectures. 
In contrast, the Copy-Paste method yields inconsistent results. While it provided gains for ViT-S, it caused a notable performance regression in the EfficientNetV2-L model, dropping accuracy from 0.5943 to 0.5721 and F1-Score from 0.7672 to 0.7463. 
This volatility suggests that while Copy-Paste increases dataset size, it introduces distributional noise that can outweigh the benefits of added diversity.

We attribute this inconsistency to the specific semantic and physical constraints of photovoltaic cells, which standard augmentation ignores:

\begin{enumerate}
    \item \textbf{Violation of Location Dependence:} Unlike generic objects, PV defects are structurally constrained by module physics. \cite{kajari2011spatial} demonstrated that cracks are not stochastic, finding that 50\% align parallel to busbars due to stress anisotropy. Similarly, \cite{heimann2014investigations} showed that contact failures are strictly bound to the electromechanical interface of the busbars. Randomly pasting these defects ignores these constraints, generating physically impossible samples that confuse the model.
    
    \item \textbf{Texture and Illumination Discontinuities:} EL images exhibit specific local illumination and grain textures. Naive pasting creates sharp, artificial discontinuities that do not exist in real manufacturing defects.
\end{enumerate}

\begin{figure}[htb!]
    \centering
    \captionsetup[subfigure]{font=scriptsize, skip=1pt}
    
    \begin{subfigure}{0.23\linewidth}
        \centering
        \includegraphics[width=\linewidth]{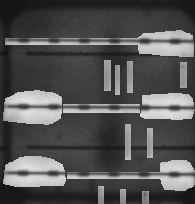}
    \end{subfigure}
    \begin{subfigure}{0.24\linewidth}
        \centering
        \includegraphics[width=\linewidth]{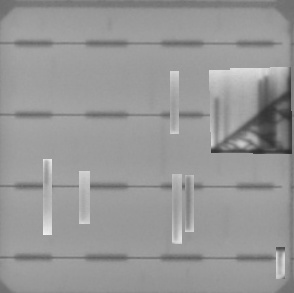}
    \end{subfigure}
    \begin{subfigure}{0.235\linewidth}
        \centering
        \includegraphics[width=\linewidth]{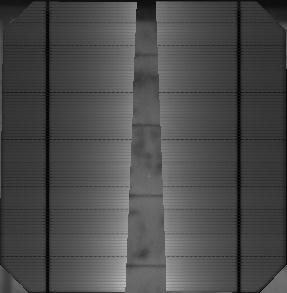}

    \end{subfigure}
    \begin{subfigure}{0.241\linewidth}
        \centering
        \includegraphics[width=\linewidth]{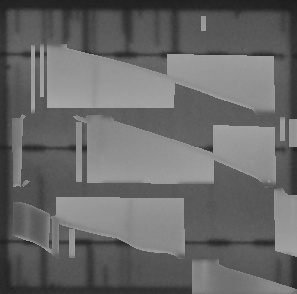}

    \end{subfigure}
    \caption{\textbf{\textit{Visual samples generated via Copy-Paste augmentation}}. The images exhibit visible texture discontinuities and artificial boundaries where defects were superimposed}
    \label{fig:copy_paste_samples}
\end{figure}

Figure~\ref{fig:copy_paste_samples} illustrates these artifacts, showing how Copy-Paste can generate physically implausible samples. In contrast, GDI is a \textit{subtractive} rather than \textit{additive} process. By strictly preserving the valid, original context of the defects that remain in the image, GDI avoids generating semantic contradictions, providing a cleaner and more physically consistent training signal.

\begin{table}[ht]
\centering
\caption{\textbf{\textit{Model Performance with Data Augmentation}}. The table compares the Baseline performance against Copy-Paste augmentation and our proposed GDI method on the full dataset. While Copy-Paste degrades performance for EfficientNetV2-L, GDI consistently improves metrics across all architectures.}

\label{tab:augmentation_results}
\resizebox{\textwidth}{!}{%
\begin{tabular}{lcccccc}
\hline
\multirow{2}{*}{Model} 
& \multicolumn{2}{c}{Baseline} 
& \multicolumn{2}{c}{Copy-Paste} 
& \multicolumn{2}{c}{GDI} \\
\cline{2-7}
& Accuracy & F1-Score 
& Accuracy & F1-Score 
& Accuracy & F1-Score \\
\hline
EfficientNetV2-L 
& 0.5943 & 0.7672 
& 0.5721 & 0.7463 
& \textbf{0.6046} & \textbf{0.7744} \\

ViT-S 
& 0.5630 & 0.7349 
& 0.5725 & 0.7501 
& \textbf{0.5832} & \textbf{0.7605} \\

ViT-L 
& 0.5336 & 0.7176 
& 0.5313 & 0.7191 
& \textbf{0.5481} & \textbf{0.7385} \\
\hline
\end{tabular}%
}
\end{table}

\subsection{Thresholding and Performance Gains}
\label{sec:robustness}

All experiments reported so far use per-class thresholds selected to maximize the macro F1-score. To assess whether the observed improvements generalize beyond a particular weight initialization or threshold choice, we repeated the 20\% data-split experiment for all three architectures across four random seeds (${24, 42, 67, 76}$) and evaluated every run at a \emph{fixed} default decision threshold of $\tau = 0.5$. This protocol is deliberately conservative: it applies no per-class threshold tuning and therefore uses no information whatsoever from the test set, making any threshold-selection bias structurally impossible.

Table~\ref{tab:robustness_seeds} reports the mean and standard deviation of Zero-One Accuracy and Macro F1 over the four seeds. Under this fixed-threshold protocol, GDI improves the mean Accuracy and mean Macro F1 for \emph{all three} architectures. The effect is also consistent at the level of individual runs: in a paired comparison (baseline versus GDI on the same seed), GDI improves Accuracy in 11 of 12 runs and Macro F1 in 10 of 12 runs as shown in Table~\ref{tab:robustness_paired}. Importantly, there is \emph{no} seed–architecture pair for which GDI causes both Accuracy and Macro F1 to decrease simultaneously; whenever one metric shows a slight reduction, the other is improved. The few exceptions are small in magnitude and fall within the observed seed-to-seed variance. Because these results require no test-set threshold tuning, they confirm that the benefit of GDI is a systematic property of the augmentation and is not an artifact of the threshold-optimization procedure used in our main experiments.

\begin{table}[h!]
\centering
\caption{\textbf{\textit{Multi-seed robustness study on the 20\% data split.}}
Mean $\pm$ standard deviation over four random seeds ($\{24, 42, 67, 76\}$),
evaluated at a fixed default threshold ($\tau = 0.5$) that uses no test-set
information. GDI improves the mean of both metrics for every architecture. Best
result within each architecture is shown in bold.}
\label{tab:robustness_seeds}
\resizebox{\textwidth}{!}{%
\begin{tabular}{llcc}
\toprule
\textbf{Architecture} & \textbf{Method} & \textbf{Accuracy (mean $\pm$ std)} & \textbf{Macro F1 (mean $\pm$ std)} \\
\midrule
\multirow{2}{*}{EfficientNetV2-L}
 & Baseline & $0.3439 \pm 0.0072$ & $0.5743 \pm 0.0168$ \\
 & GDI      & $\mathbf{0.3872 \pm 0.0318}$ & $\mathbf{0.6006 \pm 0.0057}$ \\
\midrule
\multirow{2}{*}{ViT-S}
 & Baseline & $0.3467 \pm 0.0123$ & $0.5710 \pm 0.0057$ \\
 & GDI      & $\mathbf{0.3883 \pm 0.0111}$ & $\mathbf{0.5965 \pm 0.0119}$ \\
\midrule
\multirow{2}{*}{ViT-L}
 & Baseline & $0.3083 \pm 0.0196$ & $0.5430 \pm 0.0199$ \\
 & GDI      & $\mathbf{0.3379 \pm 0.0217}$ & $\mathbf{0.5572 \pm 0.0140}$ \\
\bottomrule
\end{tabular}%
}
\end{table}

\begin{table}[h!]
\centering
\caption{\textbf{\textit{Paired per-seed comparison on the 20\% split
(fixed threshold $\tau = 0.5$).}} Each row compares baseline and GDI trained
with the \emph{same} seed; $\Delta_{\%}$ is the relative GDI improvement over
the paired baseline. This paired view controls for initialization variance.}
\label{tab:robustness_paired}
\resizebox{0.75\textwidth}{!}{%
\begin{tabular}{llcccc}
\toprule
\textbf{Architecture} & \textbf{Seed} & \textbf{$\Delta_{\%}$ Acc} & \textbf{Acc $\uparrow$} & \textbf{$\Delta_{\%}$ F1} & \textbf{F1 $\uparrow$} \\
\midrule
\multirow{4}{*}{EfficientNetV2-L}
 & 24 & $+14.5\%$ & \checkmark & $+4.9\%$  & \checkmark \\
 & 42 & $-1.0\%$  & --         & $+4.9\%$  & \checkmark \\
 & 67 & $+25.5\%$ & \checkmark & $+0.4\%$  & \checkmark \\
 & 76 & $+11.9\%$ & \checkmark & $+8.4\%$  & \checkmark \\
\midrule
\multirow{4}{*}{ViT-S}
 & 24 & $+15.0\%$ & \checkmark & $+2.6\%$  & \checkmark \\
 & 42 & $+12.0\%$ & \checkmark & $+3.5\%$  & \checkmark \\
 & 67 & $+2.7\%$  & \checkmark & $+7.3\%$  & \checkmark \\
 & 76 & $+19.0\%$ & \checkmark & $+4.6\%$  & \checkmark \\
\midrule
\multirow{4}{*}{ViT-L}
 & 24 & $+5.7\%$  & \checkmark & $+2.8\%$  & \checkmark \\
 & 42 & $+9.5\%$  & \checkmark & $-1.8\%$  & -- \\
 & 67 & $+8.9\%$  & \checkmark & $-1.5\%$  & -- \\
 & 76 & $+14.5\%$ & \checkmark & $+11.7\%$ & \checkmark \\
\bottomrule
\end{tabular}%
}
\end{table}

\section{Discussion}
\label{sec:discussion}

The experimental results strongly validate our hypothesis that isolating defects provides a superior training signal for classification models. Here, we interpret these findings and discuss their broader implications.

\subsection{Interpretation of Performance Gains}
The consistent performance improvement from GDI can be attributed to the \textbf{elimination of learning ambiguity}. In a standard multi-label setting, a classifier is presented with an image containing multiple defect patterns and is tasked with associating this complex visual input with multiple output labels. This can lead to the model learning spurious correlations or struggling to disentangle which visual features correspond to which specific defect class. 

Our GDI method helps resolve this ambiguity. By generating images where a single, clear defect pattern is present, we create a direct and clean mapping between visual evidence and its categorical label. The model is less ``confused'' by co-occurring defects and can learn a more robust and precise representation for each individual defect class. This culminates in a superior overall F1-Score.

\subsection{The Value Proposition of Annotation Repurposing}
A critical aspect of our methodology is its reliance on pixel-level segmentation masks, which are known to be labor-intensive and expensive to create \citep{fioresi2022automated}. However, this work should not be viewed as a proposal to create new segmentation datasets for classification tasks. Instead, we position GDI as a powerful technique to \textbf{maximize the return on investment for existing, high-value annotated assets}. Many research institutions and industrial quality control teams already possess such datasets. GDI provides a novel framework to repurpose this rich data to solve a different, highly practical problem.

Furthermore, this approach enables the development of lightweight and fast production-ready systems. While a segmentation model can provide detailed defect maps, such models are often computationally heavy and too slow for real-time inference on a high-throughput manufacturing line \citep{hooper2023case}. Our method allows the knowledge from these dense annotations to be distilled into a much faster classification model, which can rapidly screen and bin cells, a critical industrial requirement. GDI thus acts as a bridge, transferring detailed spatial knowledge from an offline dataset into an efficient online-capable classifier.

\begin{figure}[htbp]
    \centering
    \captionsetup[subfigure]{font=scriptsize, skip=1pt}

    \begin{subfigure}{0.24\linewidth}
        \centering
        \includegraphics[width=\linewidth]{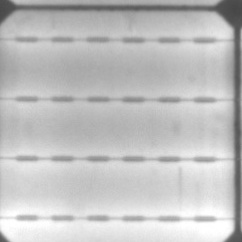}
        \caption[]{Original}
        \label{fig:fail_a_og}
    \end{subfigure}
    \begin{subfigure}{0.24\linewidth}
        \centering
        \includegraphics[width=\linewidth]{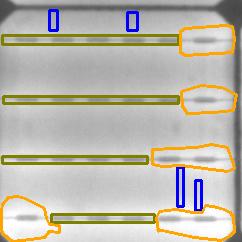}
        \caption[]{Boundaries}
        \label{fig:fail_b_bound}
    \end{subfigure}
    \begin{subfigure}{0.24\linewidth}
        \centering
        \includegraphics[width=\linewidth]{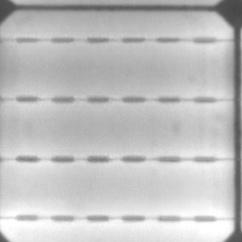}
        \caption[]{Isolated Defect}
        \label{fig:fail_c_isolated}
    \end{subfigure}
    \begin{subfigure}{0.24\linewidth}
        \centering
        \includegraphics[width=\linewidth]{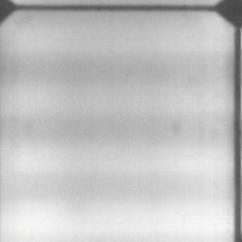}
        \caption[]{\textit{No\_Defect} (Failure)}
        \label{fig:fail_d_no_defect}
    \end{subfigure}

    \caption{\textbf{\textit{An example of a GDI failure case.}} When multiple defect annotations completely occlude a key structural feature, such as all the grid lines of the cell (b), the inpainting model lacks the context to reconstruct it. While it can successfully isolate a single defect - in this instance, the defect originally marked with an orange annotation boundary - it fails to restore the lines when generating the \textit{No\_Defect} sample (d), resulting in a visually incomplete image.}
    \label{fig:gdi_failure_case}
\end{figure}

\subsection{Limitations and Future Work}
While GDI proves highly effective, it is not without limitations. The process is contingent on the quality of the initial segmentation masks and can occasionally produce suboptimal results. A key failure mode occurs when multiple defect annotations collectively obscure a primary structural element of the cell. As illustrated in Figure~\ref{fig:gdi_failure_case}, the cell's grid lines are entirely covered by the combined defect masks. When generating the \textit{No\_Defect} sample, the inpainting model lacks any visible reference points for these lines and consequently fails to reconstruct them, resulting in a blurry and structurally incomplete image. However, it is important to note that this is an edge case. The overwhelmingly positive impact on model performance, as demonstrated on the unmodified test set, indicates that the net benefit of providing cleaner, unambiguous training data far outweighs the impact of these rare generative failures.

To address this specific limitation, a promising avenue for future research would be to develop a domain-specific inpainting model. Unlike a general-purpose model like LaMa \citep{suvorov2022resolution}, this network could be pre-trained on the consistent topology of PV cells to be structurally aware of features like grid lines. Such an approach might mitigate the primary generative failure mode and further enhance the robustness and quality of the GDI pipeline.

\section{Conclusion}
\label{sec:conclusion}

In this work, we addressed the challenge of multi-label defect classification in PV cell EL images, where the co-occurrence of multiple defects complicates model training and hinders feature disentanglement. We introduced Generative Defect Isolation (GDI), a data-centric augmentation strategy that leverages ground-truth segmentation masks to generate high-quality, single-defect training samples through neural inpainting. Extensive experiments across multiple architectures (ViT-S, ViT-L, EfficientNetV2-L) demonstrate that GDI consistently outperforms both standard baselines and prior works, establishing a new state-of-the-art with a top Zero-One Accuracy of 0.6046 and Macro F1-Score of 0.7744. The gains are most pronounced in low-data scenarios, where providing clean, unambiguous training examples proves particularly effective in mitigating the impact of data scarcity. The benefit of resolving learning ambiguity is further evidenced by a 26\% reduction in co-occurring classification errors and F1-Score improvements of up to 63.6\% for rare defect classes. Collectively, these results establish GDI as a reliable and effective method for maximizing the value of existing segmentation datasets and advancing the state of multi-label defect classification in photovoltaic inspection.

\section*{CRediT authorship contribution statement}

\textbf{Abdul Mueez:} Writing -- review \& editing, Writing -- original draft,
Visualization, Validation, Software, Methodology, Investigation, Formal
analysis, Data curation, Conceptualization.

\textbf{Yogesh S. Rawat:} Writing -- review \& editing, Visualization,
Supervision, Project administration.

\textbf{Shruti Vyas:} Writing -- review \& editing, Supervision, Resources,
Project administration, Funding acquisition, Formal analysis.

\section*{Declaration of competing interest}

The authors declare that they have no known competing financial
interests or personal relationships that could have appeared to influence
the work reported in this paper.

\section*{Data availability}

The original UCF-EL-Defect dataset used in this study is publicly
available. The source code for our pipeline and the generated synthetic
images can be accessed via our project page at
\url{https://github.com/mueez-overflow/generative-defect-isolation}.

\bibliographystyle{elsarticle-num} 
\bibliography{references}

\end{document}